\pdfoutput=1

\documentclass[11pt]{article}

\PassOptionsToPackage{table}{xcolor}

\usepackage{times}
\usepackage{latexsym}

\usepackage[T1]{fontenc}

\usepackage[utf8]{inputenc}

\usepackage{microtype}

\usepackage{inconsolata}

\usepackage{graphicx}

\usepackage{amsmath,amsfonts,amssymb}
\usepackage{bm}
\usepackage{soul}
\usepackage{array}
\usepackage{float}
\usepackage{dsfont}
\usepackage{pifont}
\usepackage{amsmath}
\usepackage{amssymb}
\usepackage{amsfonts}
\usepackage{graphicx}
\usepackage{pdfpages}
\usepackage{enumitem}
\usepackage{booktabs}
\usepackage{multirow}
\usepackage{multicol}
\usepackage{fontawesome5}
\usepackage[most]{tcolorbox}

\usepackage[final]{acl}

\usepackage{tabularx}
\usepackage[figuresright]{rotating}
\usepackage{tablefootnote}

\DeclareSymbolFont{extraup}{U}{zavm}{m}{n}
\DeclareMathSymbol{\varheart}{\mathalpha}{extraup}{86}
\DeclareMathSymbol{\vardiamond}{\mathalpha}{extraup}{87}

\usepackage{mdframed}
\mdfdefinestyle{MyQuoteFrame}{%
linecolor=black,
outerlinewidth=1em,
innerleftmargin=0.5em,
backgroundcolor=purewhite}

\usepackage[vlined]{algorithm2e}

\definecolor{yellow}{HTML}{F6BD60}
\definecolor{white}{HTML}{FFE0C1}
\definecolor{pink}{HTML}{F5CAC3}
\definecolor{tale}{HTML}{84A59D}
\definecolor{red}{HTML}{F28080}
\definecolor{orange}{HTML}{FF7F00}
\definecolor{green1}{HTML}{72C3A3}
\definecolor{green2}{HTML}{70B48F}
\definecolor{orange}{HTML}{FE8019}
\definecolor{grey}{HTML}{EBDBB2}
\definecolor{brain}{HTML}{FFABBE}
\definecolor{blue}{HTML}{076678}
\definecolor{purple}{HTML}{5861AC}
\definecolor{narrative}{HTML}{458588}
\definecolor{white2}{HTML}{F8F5E9}
\definecolor{tablewhite}{HTML}{E4E0E1}
\definecolor{purewhite}{HTML}{FFFFFF}

\newcolumntype{P}[1]{>{\centering\arraybackslash}p{#1}}

\newcommand{\nkb}{{\fontfamily{qag}\selectfont{\small Enigma}}}
\newcommand{\nkbt}{{\fontfamily{qag}\selectfont{\small\nkb$^{\textbf{\texttt{T}}}$}}}
\newcommand{\nkbp}{{\fontfamily{qag}\selectfont{\small\nkb$^{\textbf{\texttt{P}}}$}}}

\newcommand{\ours}{{\fontfamily{qag}\selectfont{\small{EnigmaToM}}}}
\newcommand{\ourstitle}{{\fontfamily{qag}\selectfont{{EnigmaToM}}}}
\newcommand{\ourssectiontitle}{{\fontfamily{qag}\fontsize{11}{6}\selectfont{{EnigmaToM}}}}
\newcommand{\nkbsectiontitle}{{\fontfamily{qag}\fontsize{11}{6}\selectfont{Enigma}}}

\newcommand{\tabbetter}[1]{\colorbox{green1}{#1}}
\newcommand{\tabworse}[1]{\colorbox{red}{#1}}

\newcommand{\tabsame}[1]{\colorbox{yellow}{#1}}

\newcommand{\vllm}{\ensuremath{%
    \mathchoice{\includegraphics[height=1.4ex]{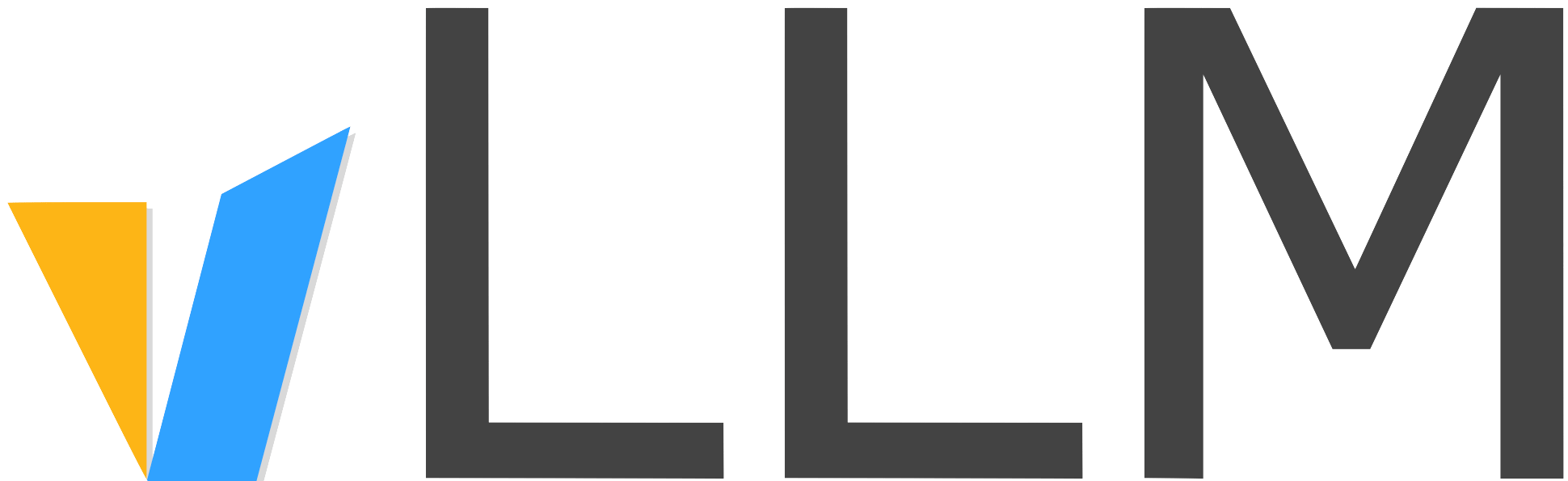}}
    {\includegraphics[height=1.4ex]{./figures/vllm.png}}
    {\includegraphics[height=1.4ex]{./figures/vllm.png}}
    {\includegraphics[height=1ex]{./figures/vllm.png}}
}}
\newcommand{\improved}{\ensuremath{%
    \mathchoice{\includegraphics[height=1.4ex]{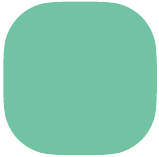}}
    {\includegraphics[height=1.4ex]{./figures/improved.pdf}}
    {\includegraphics[height=1.4ex]{./figures/improved.pdf}}
    {\includegraphics[height=1ex]{./figures/improved.pdf}}
}}
\newcommand{\unchanged}{\ensuremath{%
    \mathchoice{\includegraphics[height=1.4ex]{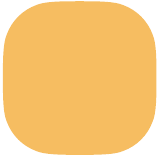}}
    {\includegraphics[height=1.4ex]{./figures/unchanged.pdf}}
    {\includegraphics[height=1.4ex]{./figures/unchanged.pdf}}
    {\includegraphics[height=1ex]{./figures/unchanged.pdf}}
}}
\newcommand{\worse}{\ensuremath{%
    \mathchoice{\includegraphics[height=1.4ex]{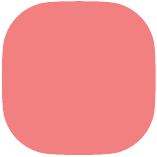}}
    {\includegraphics[height=1.4ex]{./figures/worse.pdf}}
    {\includegraphics[height=1.4ex]{./figures/worse.pdf}}
    {\includegraphics[height=1ex]{./figures/worse.pdf}}
}}

\title{\ourstitle: Improve LLMs' Theory-of-Mind Reasoning Capabilities with Neural Knowledge Base of Entity States}

\author{
    Hainiu Xu$^{\clubsuit}$ \quad 
    Siya Qi$^{\clubsuit}$ \quad 
    Jiazheng Li$^{\clubsuit}$ \quad
    Yuxiang Zhou$^{\clubsuit, \spadesuit}$ 
    \\
    \textbf{Jinhua Du}$^{\heartsuit}$ \quad
    \textbf{Caroline Catmur}$^{\clubsuit}$ \quad
    \textbf{Yulan He}$^{\clubsuit,\diamondsuit}$ \\[0.5em]
    $^\clubsuit$King's College London \quad\quad 
    $^\heartsuit$Huawei London Research Centre \\[0.25em]
    $^\diamondsuit$The Alan Turing Institute \quad\quad 
    $^\spadesuit$Queen Mary University of London \\[0.25em]
    {\tt \{hainiu.xu, yulan.he\}@kcl.ac.uk} \\
}

\begin{document}
\maketitle
\begin{abstract}
    Theory-of-Mind (ToM), the ability to infer others' perceptions and mental states, is fundamental to human interaction but remains challenging for Large Language Models (LLMs). 
    While existing ToM reasoning methods show promise with reasoning via perceptual perspective-taking, they often rely excessively on off-the-shelf LLMs, reducing their efficiency and limiting their applicability to high-order ToM reasoning. 
    To address these issues, we present \ours, a novel neuro-symbolic framework that enhances ToM reasoning by integrating a Neural Knowledge Base of entity states (\nkb) for (1) a psychology-inspired \textit{iterative masking} mechanism that facilitates accurate perspective-taking and (2) \textit{knowledge injection} that elicits key entity information. 
    \nkb~generates structured knowledge of entity states to build spatial scene graphs for belief tracking across various ToM orders and enrich events with fine-grained entity state details. 
    Experimental results on ToMi, HiToM, and FANToM benchmarks show that \ours~significantly improves ToM reasoning across LLMs of varying sizes, particularly excelling in high-order reasoning scenarios\footnote{The neural knowledge base \nkb~can be downloaded via \url{https://huggingface.co/SeacowX/Enigma}. Code and data are available at \url{https://github.com/seacowx/EnigmaToM}.}.
\end{abstract}

\section{Introduction}
Theory-of-Mind (ToM), the ability to understand that others have perceptions and mental states different from one's own, is fundamental to effective communication and social interaction \cite{premack1978does, apperly2010mindreaders}. 
ToM reasoning can be first-order, involving the understanding of another's mental state, or higher-order, requiring recursive thinking about others' beliefs. Higher-order ToM reasoning is particularly vital in real-world contexts such as negotiation \cite{de2017negotiating}.
As Large Language Models (LLMs) become increasingly sophisticated in imitating human interactions, a plethora of studies have investigated LLMs' abilities to conduct ToM reasoning.
While early studies show that LLMs exhibit traces of ToM capabilities \cite{bubeck2023sparks, kosinski2023theory}, follow-up works impugn the robustness of such capabilities by showing that LLMs' ToM reasoning is often superficial \cite{sap-etal-2022-neural, ullman2023large, shapira-etal-2024-clever}.
\begin{figure} [t!]
    \centering
    \includegraphics[width=\columnwidth]{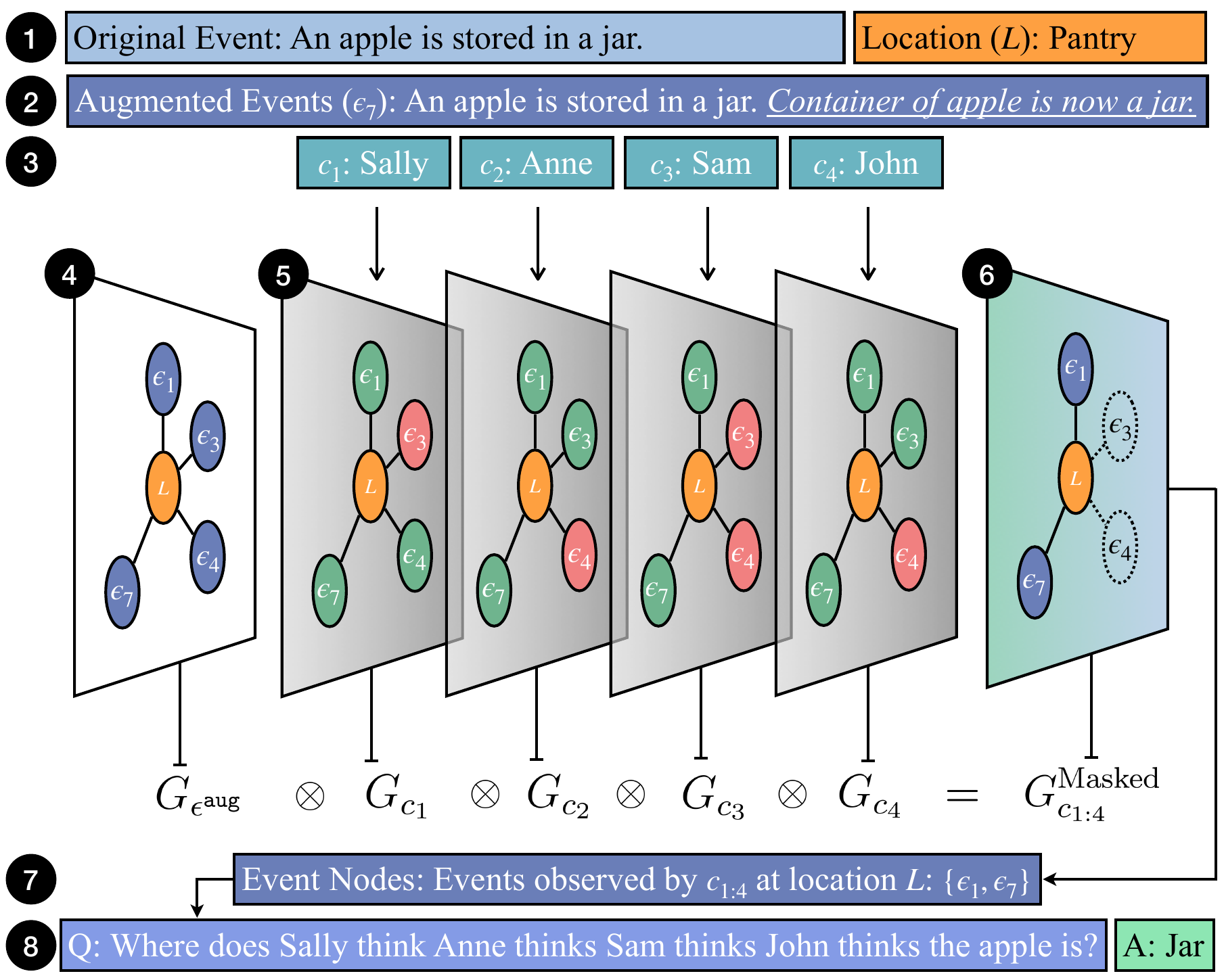}
    \caption{Example use-case of \ours~framework in fourth-order ToM reasoning. An \textit{event} (\textcircled{\raisebox{-0.3pt} {\scriptsize1}}) is enriched by adding information about entity-of-interests (\textit{italic text} in \textcircled{\raisebox{-0.3pt} {\scriptsize2}}) derived from \nkb. Characters (\textcircled{\raisebox{-0.3pt} {\scriptsize3}}) are extracted using an off-the-shelf NER model. Spatial scene graphs (\textcircled{\raisebox{-0.3pt} {\scriptsize4}} and \textcircled{\raisebox{-0.3pt} {\scriptsize5}}) are constructed for perspective-taking through a masking mechanism (\textcircled{\raisebox{-0.3pt} {\scriptsize5}} $\rightarrow$ \textcircled{\raisebox{-0.3pt} {\scriptsize6}}). Event nodes are retrieved to construct character-centric event sequence (\textcircled{\raisebox{-0.3pt} {\scriptsize7}}), which is used for the final QA (\textcircled{\raisebox{-0.3pt} {\scriptsize8}}).
    }
    \label{fig:figure1}
    \vspace{-1em}
\end{figure}


A vital prerequisite for human ToM reasoning is \textit{perceptual perspective-taking} (referred to as "perspective-taking" thereafter), which is the process of inferring the perception of other characters \cite{davis1983measuring, harwood2006conflicting}. 
In the case of ToM reasoning with LLMs, perspective-taking alleviates the reasoning burden of LLMs by identifying events that are observable by a given character and removing unobservable ones.

Centered around perspective-taking, numerous methods have been proposed. SimulatedToM \citep{wilf-etal-2024-think} and Discrete World Models (DWM) \citep{huang-etal-2024-notion} perform perspective-taking by directly prompting LLMs. While one may appreciate these methods' simplicity, the quality of perspective-taking is largely dependent on the capability of LLMs. SymbolicToM, TimeToM, and PerceptToM took a neuro-symbolic approach. TimeToM \citep{hou-etal-2024-timetom} and PerceptToM \citep{jung-etal-2024-perceptions} utilize temporal and perceptual information of events to derive characters' perception by extracting common timestamps or perceived characters. However, accurately extracting perceived timestamps or perceivers becomes difficult as the length or complexity of the event trajectory increases. The most relevant work to ours is SymbolicToM, where perspective-taking is conducted by maintaining multiple belief graphs \cite{sclar-etal-2023-minding}. 
However, SymbolicToM constructs belief graphs using less powerful models including WANLI \cite{liu-etal-2022-wanli} and OpenIE \cite{stanovsky-etal-2018-supervised}, limiting its generalizability to ToM tasks that involve complicated events. Further, as noted by \citet{sclar-etal-2023-minding}, SymbolicToM lacks efficiency as the depth of ToM reasoning increases (see \S\ref{sec:framework_efficiency} for analysis).

Given the need for accurate and efficient perspective-taking in ToM reasoning, we introduce \textbf{En}t\textbf{i}ty-\textbf{G}uided \textbf{Ma}sking (\ours), a neuro-symbolic framework enhancing LLMs' ToM reasoning (Figure~\ref{fig:figure1}).
Perspective-taking relies on reasoning about event implications, where information about the states of key entities is crucial \cite{zhang-etal-2023-causal}. \ours~employs a Neural Knowledge Base (\nkb) to generate structured entity-state information (\S\ref{sec:framework_nkb}).
This entity-state information supports spatial scene graph construction for perspective-taking (\S\ref{sec:framework_pt}) and event elicitation through knowledge injection (\S\ref{sec:framework_ki}).
Experiment results show that \ours~improves the ToM reasoning capabilities of a range of LLMs. Furthermore, the iterative masking mechanism, grounded by theories from psychology \cite{arslan2017five}, guarantees the efficacy of \ours~across ToM reasoning of varying orders.

We summarize our contributions as follows:
\begin{itemize}[leftmargin=6mm, noitemsep]
    \item[1. ] We introduce \ours, a neuro-symbolic framework for ToM reasoning that leverages a Neural Knowledge Base of Entity States to improve LLMs' ToM reasoning capabilities.
    \item[2. ] Through the iterative masking mechanism, \ours~conducts effective perspective-taking while greatly reducing the number of character belief graphs that need to be tracked, thereby improving the efficiency in high-order ToM reasoning. 
    \item[3. ] \ours~improves LLMs' ToM reasoning, especially for higher-order cases. Analysis show that \ours~improves LLMs' ToM reasoning ability up to the fourth order.
\end{itemize}

\section{Related Work}
\paragraph{Knowledge Base of Commonsense Knowledge in Natural Language}
Efforts to construct commonsense knowledge bases have a long history.
Early work includes CyC, ConceptNet, and DBPedia \cite{lenat1995cyc, Liu2004ConceptNetA, lehmann2015dbpedia}. \citet{rashkin-etal-2018-event2mind} introduced Event2Mind, an event-based knowledge graph that captures characters' intentions and reactions. Subsequently, \citet{sap2019atomic} introduced ATOMIC, a commonsense knowledge graph that models if-then relationships for simple events. 
To explore more complex events, \citet{tandon-etal-2020-dataset} introduced OpenPI, a dataset for entity state tracking in procedures. OpenPI was extended to OpenPI2.0 by introducing entity saliency scores and entity canonicalization \citep{zhang-etal-2024-openpi2}. 
Parallel efforts have developed neural models, including a GRU-based encoder-decoder model for Event2Mind \citep{rashkin-etal-2018-event2mind}, a decoder-only Transformer called COMET for ConceptNet and ATOMIC \cite{bosselut-etal-2019-comet}, and fine-tuned GPT-2 for OpenPI \citep{tandon-etal-2020-dataset}. 

\begin{figure*}
    \centering
    \includegraphics[width=\textwidth]{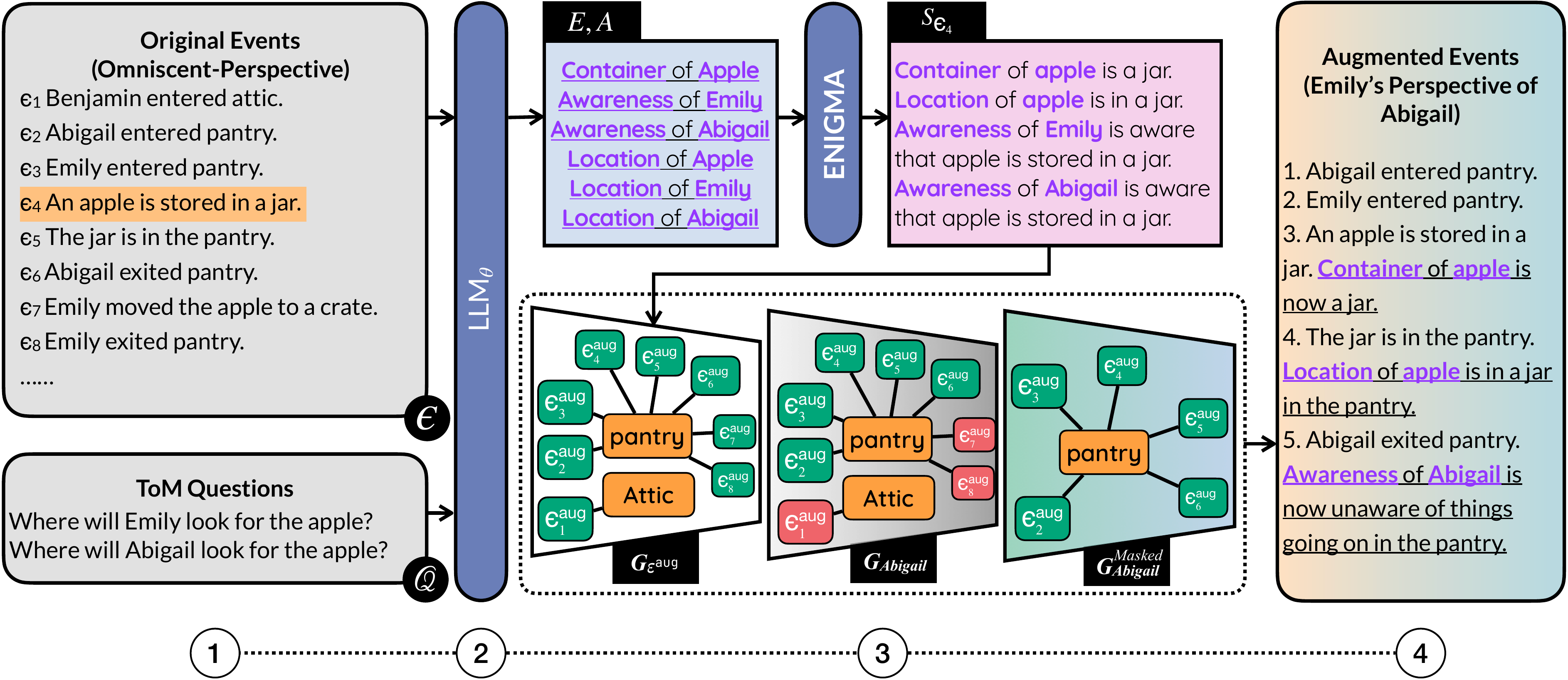}
    \caption{An overview of the \ours~framework. In the graphs shown at bottom of \raisebox{.5pt}{\textcircled{\raisebox{-0.pt} {\scriptsize3}}}, $\improved$ nodes denotes observed events while $\worse$ nodes denotes unobserved events. See detailed explanations in \S\ref{para:ours_overview}.}
    \label{fig:enigma_pipeline}
    \vspace{-1em}
\end{figure*}

\paragraph{Benchmarking LLMs' ToM Reasoning Capabilities}
Many ToM benchmarks are inspired by the False Beliefs test \cite{wimmer1983beliefs}, including event-based benchmarks such as ToMi \cite{le-etal-2019-revisiting}, HiToM \cite{wu-etal-2023-hi}, BigToM \cite{gandhi2024understanding}, and OpenToM \cite{xu-etal-2024-opentom}, and dialogue-based datasets such as FANToM \cite{kim-etal-2023-fantom}. Based on the Smarties Test \cite{gopnik1988children}, Adv-CSFB \cite{shapira-etal-2024-clever} and ToMChallenges \cite{ma-etal-2023-tomchallenges} assess LLMs' ability to reason about unexpected contents and unexpected transfers. ToMBench \cite{chen-etal-2024-tombench} and EPITOME \cite{jones2023epitome} contain a suite of ToM tasks that go beyond False Beliefs and Smarties Test. MMToM-QA extends ToM evaluation to multimodality \cite{jin-etal-2024-mmtom} and InformativeBench evaluates ToM in multi-agent settings \cite{liuautonomous}.

\paragraph{Improving LLMs' ToM Reasoning Capabilities}
Methods for improving LLMs' ToM reasoning capabilities have focused on \textit{perspective-taking}. 
SymbolicToM conducts perspective-taking via belief graphs \cite{sclar-etal-2023-minding}. 
SimulatedToM \cite{wilf-etal-2024-think} and DWM \cite{huang-etal-2024-notion} conduct perspective-taking by prompting. DWM additionally prompts LLMs to infer the world state after a group of events.
TimeToM utilizes the temporal order of events to conduct perspective-taking \cite{hou-etal-2024-timetom}. 
PerceptToM does perspective-taking by prompting LLMs to infer perceivers of each event \cite{jung-etal-2024-perceptions}. 
For multimodal ToM, methods like NIPE and BIP-ALM leverage Bayesian Inverse Planning \cite{yingneuro, jin-etal-2024-mmtom}, with environments (e.g. 2D grids or videos) providing strong perspective-taking signals through observable trajectories. 

\section{The \ourssectiontitle~Framework}
\label{sec:framework}

Before presenting the \ours~framework, we define the general setup of ToM reasoning tasks.

\paragraph{ToM Task Setup}
\label{sec:framework_setup}
We focus on the widely studied ToM task of \textit{reasoning about false beliefs} \cite{wimmer1983beliefs}, which is typically formulated as QA tasks. Formally, given a context consisting of a sequence of events, $\mathcal{E} = \{\epsilon_i\}_{i=1}^{n}$, which involves multiple characters, $\mathcal{C}=\{c_j\}_{j=1}^{m}$, and a query regarding the belief of a particular character, $q_{c}, c \in \mathcal{C}$, the goal is to derive the most likely belief, $b_{c}$, from all potential beliefs, $\mathcal{B}_{c}$:
\begin{equation}
b^*_{c} = \arg\max_{b \in \mathcal{B}_{c}} \mathbb{P}(b | \mathcal{E}, q_{c}, c \in \mathcal{C})
\end{equation}
Further, $\mathcal{E}$ can be concise events as seen in the ToMi dataset \cite{le-etal-2019-revisiting} or utterances as seen in the FANToM dataset \cite{kim-etal-2023-fantom}. 
Beyond directly querying a character’s beliefs about the environment, one can also probe their beliefs regarding other characters’ perceptions, thereby enabling the assessment of higher-order ToM reasoning.

\vspace{-0.1em}
\paragraph{The \ourssectiontitle~Framework}
\label{para:ours_overview}
Figure~\ref{fig:enigma_pipeline} provides an overview of our framework, we use circled number (\raisebox{.0pt}{\textcircled{\raisebox{-0.pt} {\scriptsize$\mathbb{N}$}}}) to refer to components in the figure.
At the core of \ours~is a Neural Knowledge Base (NKB) of Entity States (\nkb). 
Given a sequence of events (\raisebox{.5pt}{\textcircled{\raisebox{-0.pt} {\scriptsize1}}}.$\mathcal{E}$) and the corresponding questions (\raisebox{.5pt}{\textcircled{\raisebox{-0.pt} {\scriptsize1}}}.$\mathcal{Q}$), \ours~first leverages a chosen LLM (\raisebox{.5pt}{\textcircled{\raisebox{-0.pt} {\scriptsize2}}}) to identify key entities (e.g., characters and important objects) and their attributes relevant to ToM reasoning (Top left of \raisebox{.5pt}{\textcircled{\raisebox{-0.pt} {\scriptsize3}}}). 
\nkb then produces state information for these entities after each event (\raisebox{.5pt}{Top right of \textcircled{\raisebox{-0.pt} {\scriptsize3}}}, \S\ref{sec:framework_nkb}). 
With the entity state knowledge, \ours~first conducts \textit{Knowledge Injection} (referred to as "\texttt{KI}" thereafter) to enrich the original events by adding relevant fine-grained entity state details (\S\ref{sec:framework_ki}). 
Among the entity state knowledge, spatial information of characters is used to conduct perspective-taking through an \textit{Iterative Masking} mechanism (referred to as "\texttt{IM}" thereafter. Bottom of \raisebox{.5pt}{\textcircled{\raisebox{-0.pt} {\scriptsize3}}}, \S\ref{sec:framework_pt}). The modified events are provided to the LLM for final answers via zero-shot prompting (\raisebox{.5pt}{\textcircled{\raisebox{-0.pt} {\scriptsize4}}}). 
By offloading much of the ToM reasoning process to the symbolic \texttt{IM} component via perspective-taking, \ours~reduces LLMs' reasoning burden.

\subsection{The \nkb~Neural Knowledge Base}
\label{sec:framework_nkb}
NKBs such as COMET are trained on a large corpus of structured knowledge in a sequence-to-sequence manner \cite{bosselut-etal-2019-comet}. Following this approach, we fine-tuned a Llama3.1-8B \cite{dubey2024llama} model to function as our NKB \cite{llamafactory}.\footnote{See Appendix~\ref{app:training_nkbt} for details of the fine-tuning process.} For training, we used OpenPI2.0 \cite{zhang-etal-2024-openpi2}, which consists of 25,600 human-annotated entity state changes derived from WikiHow 
articles. OpenPI2.0 was selected over ATOMIC and Event2Mind as it contains more complex events and entity states. 
Alternatively, as LLMs become increasingly adept at commonsense reasoning, they can serve as an NKB of entity states via prompting \cite{hwang2021comet}. We denote the trained (\textbf{\texttt{T}}) and prompt-based (\textbf{\texttt{P}}) NKB as \nkbt~and \nkbp, respectively.

To query the NKB, we adopt an \textit{entity-attribute-guided} approach which contains two steps. In Step 1, given a sequence of events, $\mathcal{E}$, a set of ToM questions, $\mathcal{Q}$, and a chosen LLM parameterized by $\theta$, we obtain a set of entities of interest, $E = \{e\}_{i=1}^{n}$, and their corresponding attributes, $A = \{a\}_{j=1}^{m}$, by zero-shot prompting:
\begin{equation}
\label{eq:eoi_query}
E, A = \texttt{LLM}_{\theta}\Big(\rho(\mathcal{E}, \mathcal{Q})\Big)
\end{equation}
where $\rho$ denotes the prompt template (see Appendix~\ref{app:prompt_bank} for details of the prompt).
Then in Step 2, given an event, $\epsilon \in \mathcal{E}$, a set of entities of interest $E$ and their corresponding attributes $A$, we query \nkb~to retrieve the state of the entities after event $\epsilon$:
\begin{align}
\label{eq:nkb_query}
s_{\epsilon} &= \bigoplus_{i=1}^{n} \bigoplus_{j=1}^{m} \text{\nkb}(e_i, a_j, \epsilon) \notag \\
&\quad\quad\quad\quad\quad \forall \epsilon \in \mathcal{E},\ e_i \in E,\ a_j \in A
\end{align}

where $\oplus$ denotes concatenation. 

\subsection{Knowledge Injection (\texttt{KI}) with \nkb}
\label{sec:framework_ki}
In prior studies, perspective-taking was regarded as filtering out events unobserved by a given character, yielding a subset $\mathcal{E}'_{c} \subseteq \mathcal{E}$. We argue that beyond event filtering, perspective-taking should enhance LLMs' comprehension of events.
Fine-grained entity state knowledge is crucial for event reasoning \cite{zhang-etal-2023-causal} but often omitted due to reporting bias \cite{shwartz-choi-2020-neural}. To address this, we propose a knowledge injection mechanism, \texttt{KI}, that utilizes \nkb~to enrich observable events with fine-grained entity state information.
In the first step of \texttt{KI}, a chosen LLM is used to infer key entities, $E$, and attributes, $A$, based on a given sequence of events, $\mathcal{E}$, and a set of ToM questions, $\mathcal{Q}$, (Equation~\ref{eq:eoi_query}). We then query \nkb~with the recognizied entities and their attributes to obtain their state information at each event (Equation~\ref{eq:nkb_query}). We exclude spatial information of characters, $\mathcal{S}^p_{c}$, as this will be handled in the subsequent masking process (\S\ref{sec:framework_pt}). Given a sequence of events, $\mathcal{E} = \{\epsilon\}_{i=1}^n$, we augment it by injecting entity state knowledge, resulting in the sequence $\mathcal{E}^{\texttt{aug}}$:  
\begin{equation}
\label{eq:ki}
\mathcal{E}^{\texttt{aug}} = \bigoplus_{i=1}^{n} \epsilon_i \oplus \hat{s}_{\epsilon_i}, \text{ where } \hat{s}_{\epsilon_i} = s_{\epsilon_i} \setminus s^p_{c}
\end{equation}
where $\oplus$ denotes concatenation. 
As fine-grained entity state knowledge is often omitted in events due to reporting bias \cite{shwartz-choi-2020-neural}, this mechanism compensates for the lost information. 
More importantly, by providing state information of key entities, \texttt{KI} reinforces LLMs' understanding of the observed events.

\subsection{Perspective-Taking (\texttt{IM}) with \nkb}
\label{sec:framework_pt}

Studies in psychology have shown that people's beliefs about others' mental states rely only on information available to themselves\footnote{For instance, "Anne's belief about Sally's mental state" depends only on information available to Anne, i.e. events witnessed by Anne herself.} \cite{arslan2017five}.  
Building on this insight, we assume that characters interpret others' beliefs through the lens of their own mental states, which allows us to employ \textit{Iterative Masking} (\texttt{IM}) to facilitate efficient and accurate ToM reasoning across various order.

Perspective-taking with \nkb~is accomplished by constructing spatial scene graphs and performing \textit{Iterative Masking} (\texttt{IM}) using constructed graphs. Specifically, we obtain spatial information, $\mathcal{S}^p_{c}$, by querying \nkb~about the location ($attr$) of a specific character ($ent$), $c$, using Equation (\ref{eq:nkb_query}). Spatial scene graphs are constructed based on spatial information to represent the detailed locations where each event takes place as perceived by a given character. The nodes of the scene graph represent events and locations, while the edges denote the "\texttt{isin}" relationship, specifying the location where each event takes place.

During \texttt{IM}, we first construct a character-oblivious spatial scene graph, $G_{\mathcal{E}^{\texttt{aug}}}$, which documents the \textit{location} of each augmented event from an omniscient perspective. 
We then construct character-centric spatial scene graphs, $G_{c}$, that capture event locations from the perspective of each character. 
We introduce a null node, $\varnothing$, which indicates that the location of the current event is unknown to the character. During \texttt{IM}, the null node serves as a "\textit{mask}" to exclude the event nodes, which are unobserved by the character, from $G_{\mathcal{E}^{\texttt{aug}}}$ (see Figure~\ref{fig:figure1} and Figure~\ref{fig:enigma_pipeline}). For high-order ToM reasoning, $G_{\mathcal{E}^{\texttt{aug}}}$ is masked sequentially by the order of characters in the belief chain\footnote{For instance, the masked spatial scene graph for "Sally's belief of Anne's mental state" is $G_{\mathcal{E}^{\texttt{aug}}} \otimes G_{\text{Sally}} \otimes G_{\text{Anne}}$.}:
\begin{equation}
\label{eq:mask}
G^{\text{masked}}_{c_{1:k}} = G_{\mathcal{E}^{\texttt{aug}}} \bigotimes_{j=1}^{k} G_{c_j}
\end{equation}
where $\otimes$ represents the masking operation, and $k$ corresponds to the ToM-order. The observable events of character $c_{1:k}$ with injected entity state knowledge can be constructed as:
\begin{equation}
\label{eq:retrieve_events}
\mathcal{E}^{\texttt{aug}}_{c_{1:k}} = V_{G^{\text{masked}}_{c_{1:k}}}^{\epsilon}
\end{equation}
where $V_{G^{\text{masked}}_{c_{1:k}}}^{\epsilon}$ represents event nodes in $G^{\text{masked}}_{c_{1:k}}$. In the case of high-order Tom reasoning,  $\mathcal{E}^{\texttt{aug}}_{c_{1:k}}$ is obtained by iteratively applying the belief of characters. As such, $\mathcal{E}^{\texttt{aug}}_{c_{1:k}}$ effectively encapsulates the beliefs of all characters in the belief chain. This allows us to transform the high-order ToM question to that of first-order.
For instance, reasoning about “Sally’s belief about Anne’s belief” without \ours~requires first inferring Sally’s perceived world state, which then serves as the basis for modeling Anne’s belief. 
With \ours, such nested dependencies and recursive reasoning are handled by the \texttt{IM} mechanism. Consequently, under $\mathcal{E}^{\texttt{aug}}_{\text{Sally}, \text{Anne}}$, deriving Sally's belief is sufficient to answer the original second-order Theory of Mind (ToM) question. Illustrative examples and further details on ToM order reduction are provided in Appendix~\ref{app:reduce_tom_question_order}.
We present illustrative examples and details of ToM order reduction in  Appendix~\ref{app:reduce_tom_question_order}.

\subsection{Efficiency of \ourssectiontitle}
\label{sec:framework_efficiency}
The \texttt{IM} mechanism of \ours~addresses the intractability of high-order ToM reasoning faced by SymbolicToM \cite{sclar-etal-2023-minding}. Due to the asymmetry of ToM modeling\footnote{For example, in second-order ToM, Anne's belief of Sally's mental state is not equivalent to Sally's belief of Anne's mental state.}, enumerating all possible mental states for characters across all ToM orders is a permutation problem.
Suppose a ToM reasoning question involves $m$ characters and the ToM order goes up to $k^{th}$-order, the worst-case complexity of constructing belief graphs in SymbolicToM is $\mathcal{O}\Big(\sum_{i=1}^k\frac{m!}{(m-i)!}\Big)$.
In contrast, \ours~constructs one spatial scene graph, $G_{\mathcal{E}^{\texttt{aug}}}$, which encapsulates omniscient spatial information, and $m$ character-centric spatial scene graphs. Hence, the worst-case complexity for constructing spatial scene graphs in \ours~is $\tilde{T}(m, k)=\mathcal{O}(m)$, which is linear with respect to the number of characters and independent of the ToM order $k$. We illustrate the difference in complexity in Appendix~\ref{app:vis_complexity}.

\begin{table}
    \centering 
    \Huge
    \resizebox{\columnwidth}{!}{
        \begin{tabular}{ c c c c c }
            \toprule
            \textbf{{Dataset}} & \textbf{{O}} & \textbf{{Unit}} & \textbf{{\#Units}} & \textbf{{\#Qs}} \\
            \midrule
            \addlinespace[0.5ex]
            ToMi\tablefootnote{We use the disambiguated ToMi \cite{sclar-etal-2023-minding} from \url{https://github.com/msclar/symbolictom}.} \cite{le-etal-2019-revisiting} & 2 & E & 9.85 & 114 \\
            HiToM \cite{wu-etal-2023-hi} & 4 & E & 26.49 & 614 \\
            FANToM \cite{kim-etal-2023-fantom} & 2 & U& 23.14 & 577 \\
            \bottomrule
        \end{tabular}
    }
    \caption{Summary of datasets. \textbf{\textit{O}}: highest ToM order tested. \textbf{\textit{Unit}}: type of event sequence. "E" for event and "U" for utterance. \textbf{\textit{\#Units}}: avg. units per sequence. \textbf{\textit{\#Qs}}: avg. number of questions per sampled subset. Examples from each dataset can be found in Appendix~\ref{app:data_examples}.}
    \label{tab:datasets}
    \vspace{-0.5em}
\end{table}

\begin{table*} [h]
    \centering
    \resizebox{\linewidth}{!}{
    \begin{tabular} {P{0.01\columnwidth} c c c c c c c c c}
        \toprule
        \addlinespace[0.25ex]
        & & \textbf{\texttt{Qwen2.5-7B}} & \textbf{\texttt{Llama3.1-8B}} & \textbf{\texttt{Gemma2-9B}} & \textbf{\texttt{Gemma2-27B}} & \textbf{\texttt{Llama3.3-70B}$^{\texttt{4bit}}$} & \textbf{\texttt{Qwen2.5-72B}$^{\texttt{4bit}}$} & \textbf{\texttt{GPT-4o}} \\
        \midrule
        \addlinespace[0.5ex]
        \multirow{8}{*}{\rotatebox[origin=c]{90}{ToMi}} & 
        Vanilla & $0.722_{\pm0.045}$ & $0.647_{\pm0.011}$ & $0.741_{\pm0.037}$ & $0.715_{\pm0.048}$ & $0.767_{\pm0.015}$ & $0.717_{\pm0.034}$ & $0.767_{\pm0.041}$ \\
        & CoT & \underline{$0.724_{\pm0.026}$} & \underline{$0.739_{\pm0.025}$} & $0.676_{\pm0.035}$ & $0.537_{\pm0.056}$ & $0.741_{\pm0.032}$ & $0.767_{\pm0.033}$ & $0.769_{\pm0.029}$ \\
        & SimToM & $0.642_{\pm0.022}$ & $0.600_{\pm0.020}$ & $0.710_{\pm0.034}$ & $0.684_{\pm0.015}$ & $0.712_{\pm0.018}$ & $0.749_{\pm0.020}$ & $0.749_{\pm0.018}$ \\
        & TimeToM & $0.567_{\pm0.024}$ & $0.630_{\pm0.019}$ & $0.681_{\pm0.028}$ & $0.587_{\pm0.036}$ & $0.739_{\pm0.021}$ & \boldmath{$0.865_{\pm0.018}$} & $0.723_{\pm0.016}$ \\
        & DWM & $0.686_{\pm0.023}$ & $0.644_{\pm0.033}$ & $0.718_{\pm0.028}$ & $0.707_{\pm0.045}$ & $0.735_{\pm0.016}$ & $0.762_{\pm0.051}$ & $0.739_{\pm0.049}$ \\
        & PerceptToM & $0.720_{\pm0.038}$ & $0.695_{\pm0.025}$ & $0.676_{\pm0.029}$ & $0.749_{\pm0.017}$ & $0.738_{\pm0.032}$ & $0.809_{\pm0.033}$ & $0.790_{\pm0.023}$ \\[0.2em]
        \cmidrule{2-9}
        \addlinespace[0.5ex]
        \rowcolor{tablewhite} & \textbf{\nkbp} & $0.706_{\pm0.044}$ & $0.738_{\pm0.056}$ & \boldmath{$0.865_{\pm0.031}$} & \boldmath{$0.833_{\pm0.018}$} & \boldmath{$0.828_{\pm0.012}$} & \underline{$0.839_{\pm0.014}$} & \boldmath{$0.847_{\pm0.030}$} \\
        \rowcolor{tablewhite} & \textbf{\nkbt} & \boldmath{$0.825_{\pm0.030}$} & \boldmath{$0.796_{\pm0.023}$} & \underline{$0.814_{\pm0.020}$} & \underline{$0.804_{\pm0.050}$} & \underline{$0.787_{\pm0.024}$} & $0.837_{\pm0.024}$ & \underline{$0.795_{\pm0.036}$} \\

        \addlinespace[0.5ex]
        \bottomrule
        \addlinespace[0.5ex]

        \multirow{8}{*}{\rotatebox[origin=c]{90}{HiToM}} & 
        Vanilla & $0.378_{\pm0.013}$ & $0.333_{\pm0.015}$ & $0.471_{\pm0.009}$ & $0.527_{\pm0.018}$ & $0.534_{\pm0.008}$ & $0.456_{\pm0.012}$ & $0.521_{\pm0.006}$ \\
        & CoT & $0.441_{\pm0.007}$ & $0.304_{\pm0.021}$ & $0.474_{\pm0.008}$ & $0.535_{\pm0.018}$ & $0.537_{\pm0.011}$ & $0.481_{\pm0.011}$ & $0.527_{\pm0.005}$ \\
        & SimToM & $0.402_{\pm0.009}$ & $0.368_{\pm0.024}$ & $0.473_{\pm0.012}$ & \underline{$0.549_{\pm0.018}$} & $0.569_{\pm0.005}$ & $0.536_{\pm0.018}$ & $0.571_{\pm0.003}$ \\
        & TimeToM & $0.316_{\pm0.010}$ & \underline{$0.462_{\pm0.013}$} & $0.302_{\pm0.012}$ & $0.302_{\pm0.013}$ & \underline{$0.623_{\pm0.006}$} & $0.415_{\pm0.013}$ & \underline{$0.633_{\pm0.008}$} \\
        & DWM & $0.444_{\pm0.020}$ & $0.367_{\pm0.019}$ & \underline{$0.485_{\pm0.012}$} & $0.488_{\pm0.018}$ & $0.564_{\pm0.010}$ & \underline{$0.560_{\pm0.009}$} & $0.580_{\pm0.018}$ \\
        & PerceptToM & $0.393_{\pm0.019}$ & $0.342_{\pm0.011}$ & $0.440_{\pm0.009}$ & $0.562_{\pm0.007}$ & $0.588_{\pm0.010}$ & $0.548_{\pm0.016}$ & $0.580_{\pm0.018}$ \\[0.2em]
        \cmidrule{2-9}
        \addlinespace[0.5ex]
        \rowcolor{tablewhite} & \textbf{\nkbp} & \boldmath{$0.508_{\pm0.012}$} & \boldmath{$0.477_{\pm0.005}$} & \boldmath{$0.555_{\pm0.010}$} & \boldmath{$0.576_{\pm0.004}$} & \boldmath{$0.696_{\pm0.007}$} & \boldmath{$0.605_{\pm0.007}$} & \boldmath{$0.733_{\pm0.017}$} \\
        \rowcolor{tablewhite} & \textbf{\nkbt} & \underline{$0.457_{\pm0.005}$} & $0.431_{\pm0.010}$ & $0.446_{\pm0.008}$ & $0.478_{\pm0.004}$ & $0.518_{\pm0.011}$ & $0.473_{\pm0.010}$ & $0.626_{\pm0.020}$\\

        \addlinespace[0.5ex]
        \bottomrule
        \addlinespace[0.5ex]

        \multirow{8}{*}{\rotatebox[origin=c]{90}{FANToM}} & 
        Vanilla & $0.400_{\pm0.015}$ & $0.429_{\pm0.022}$ & $0.485_{\pm0.016}$ & $0.553_{\pm0.011}$ & $0.486_{\pm0.022}$ & $0.532_{\pm0.025}$ & $0.476_{\pm0.020}$ \\
        & CoT & $0.398_{\pm0.014}$ & $0.438_{\pm0.014}$ & $0.470_{\pm0.019}$ & $0.556_{\pm0.007}$ & $0.494_{\pm0.028}$ & $0.521_{\pm0.024}$ & $0.453_{\pm0.014}$ \\
        & SimToM & $0.413_{\pm0.012}$ & $0.440_{\pm0.015}$ & $0.427_{\pm0.009}$ & $0.574_{\pm0.010}$ & \boldmath{$0.620_{\pm0.025}$} & $0.516_{\pm0.014}$ & $0.502_{\pm0.016}$ \\
        & TimeToM & $0.252_{\pm0.020}$ & $0.260_{\pm0.012}$ & $0.299_{\pm0.011}$ & $0.300_{\pm0.021}$ & $0.580_{\pm0.017}$ & $0.409_{\pm0.026}$ & $0.404_{\pm0.016}$ \\
        & DWM & $0.429_{\pm0.013}$ & \underline{$0.470_{\pm0.027}$} & $0.433_{\pm0.023}$ & $0.562_{\pm0.017}$ & $0.473_{\pm0.021}$ & $0.543_{\pm0.014}$ & $0.465_{\pm0.028}$ \\
        & PerceptToM & $0.408_{\pm0.023}$ & $0.407_{\pm0.026}$ & \underline{$0.504_{\pm0.006}$} & \boldmath{$0.611_{\pm0.009}$} & $0.527_{\pm0.011}$ & \underline{$0.573_{\pm0.016}$} & $0.521_{\pm0.006}$ \\[0.2em]
        \cmidrule{2-9}
        \addlinespace[0.5ex]
        \rowcolor{tablewhite} & \textbf{\nkbp} & \underline{$0.445_{\pm0.026}$} & $0.442_{\pm0.018}$ & $0.439_{\pm0.023}$ & $0.462_{\pm0.014}$ & $0.515_{\pm0.020}$ & $0.450_{\pm0.013}$ & \underline{$0.531_{\pm0.015}$} \\
        \rowcolor{tablewhite} & \textbf{\nkbt} & \boldmath{$0.487_{\pm0.018}$} & \boldmath{$0.545_{\pm0.036}$} & \boldmath{$0.530_{\pm0.012}$} & \underline{$0.582_{\pm0.028}$} & \underline{$0.610_{\pm0.021}$} & \boldmath{$0.574_{\pm0.031}$} & \boldmath{$0.553_{\pm0.011}$}\\
        \addlinespace[0.5ex]
        \bottomrule
    \end{tabular}
    }
    \caption{Main results of \ours~in comparison with existing methods on ToMi, HiToM, and FANToM datasets. Accuracy means and variances are calculated based on 5 runs, which used 5 different subsets of the corresponding dataset. The best and second best results are highlighted in \textbf{bold} and \underline{underline} respectively.}
    \label{tab:main_results}
    \vspace{-1em}
\end{table*}

\section{Experiments}
\label{sec:exp}
\ours~is evaluated on three widely used ToM benchmarks (Table~\ref{tab:datasets}) and compared against the following generic and ToM-specific methods:
\renewcommand{\thefootnote}{$\dagger$}
\begin{itemize}[leftmargin=*, noitemsep, label={}]
    \item{\textbf{CoT}} \citep{wei2022chain} boosts LLMs' reasoning capabilities by prompting LLMs to explicitly list out their reasoning process.
    \item \textbf{SimToM} \citep{wilf-etal-2024-think} conducts perspective-taking by directly querying the LLMs about the mental states of characters.
    \item \textbf{TimeToM}\footnote{Official implementation is not available at the time of experiments (Sept-Dec, 2024). We implemented this method using prompts from the corresponding paper.}   
    \citep{hou-etal-2024-timetom} leverage the temporal information of events to conduct perspective-taking. The final answer is obtained using a multi-perspective belief-solving prompt.  
    \item \textbf{DWM} \citep{huang-etal-2024-notion} conducts perspective-taking by partitioning the events into chunks and querying the LLMs about characters' mental states after each chunk.
    \item \textbf{PerceptToM}$^{\dagger}$ \citep{jung-etal-2024-perceptions} conducts perspective-taking by querying the LLMs about the characters' awareness of the events.
\end{itemize}

\renewcommand{\thefootnote}{\arabic{footnote}}
\noindent
To ensure a fair comparison with established methods, we conduct controlled experiments by controlling the format and answer space of all ToM questions. In addition, we follow a realistic setting of ToM reasoning by using only the sequence of events and ToM questions from each dataset. Auxiliary information such as character names is obtained using an off-the-shelf NER model\footnote{\url{https://huggingface.co/dslim/bert-large-NER}}.

\paragraph{Question Formatting}
We formulate ToMi as a free-form generation task where the model is instructed to choose between two possible answers. We formulate HiToM as a multiple-choice task as in the original paper \cite{wu-etal-2023-hi}. FANToM contains both free-form generation and multiple-choice questions. We follow the question formatting instructions in the original paper \cite{kim-etal-2023-fantom}. For efficient and accurate parsing of LLM responses, we follow the convention of \citep{huang-etal-2024-notion}, instructing LLMs to wrap answers within the special \texttt{<answer>} and \texttt{</answer>} tokens. 
As introduced in \S\ref{sec:framework_pt}, the recursive modeling of mental states in high-order ToM questions has been addressed by the \texttt{IM} mechanism, which allows us to transform high-order ToM questions into first-order questions. Similarly, TimeToM leverages temporal information to conduct symbolic modeling of high-order ToM \cite{hou-etal-2024-timetom}. We apply such transformation when evaluating with TimeToM and \ours~ (Appendix~\ref{app:reduce_tom_question_order}).

\renewcommand{\thefootnote}{$\ddagger$}
\paragraph{Towards Robust Evaluation}
To ensure robust evaluation, we construct 5 subsets for each dataset by sampling data points using commonly used random seeds\footnote{We use \texttt{12}, \texttt{42}, \texttt{96}, \texttt{2012}, and \texttt{2024} as random seeds.}. Each subset of ToMi and HiToM contains 100 event sequences, whereas each subset of FANToM contains 50 multi-round dialogues. The number of QA pairs in each subset is shown in Table~\ref{tab:datasets}. We report both the mean accuracy and its variance based on the 5 runs. \\

\noindent
We evaluate each method using various instruction-tuned LLMs, including Llama3.1-8B, Llama3.3-70B\footnote{Loaded in 4bit using BitsandBytes \cite{dettmers8} with weights from \url{https://huggingface.co/unsloth}.}, Qwen2.5-7B, Qwen2.5-72B$^{\ddagger}$, Gemma2-9B, Gemma2-27B, 
\renewcommand{\thefootnote}{\arabic{footnote}}
and GPT-4o \cite{dubey2024llama, qwen2.5, team2024gemma, hurst2024gpt}. To ensure reproducibility, all experiments are done using zero-shot prompting with greedy decoding and a temperature of \texttt{0}. LLM inference is carried out using $\vllm$ on 2 NVIDIA A100$^{80\texttt{GB}}$ GPUs \cite{kwon2023efficient}. 

Table~\ref{tab:main_results} shows the main results of \ours~in comparison with existing methods on ToMi, HiToM, and FANToM datasets. In general, we see that \ours~brings improvements in accuracy across all datasets and most LLMs. Specifically, \nkbp~outperforms other methods on ToMi and HiToM, while \nkbt~achieves superior performance on FANToM. \ours~is particularly effective with smaller LLMs. For instance, \nkbt~boosts Qwen2.5-7B to exceed the zero-shot performance of Qwen2.5-72B$^{\texttt{4bit}}$. Further, results from the HiToM dataset demonstrate that \ours~is particularly effective in high-order ToM reasoning. We analyze the effectiveness of \ours~in tackling high-order ToM reasoning 
in \S\ref{sec:analysis_high_order}. Moreover, results from Table~\ref{tab:main_results} show that \nkbp~performs better on event-based datasets (ToMi and HiToM) while \nkbt~is more effective on a dialogue-based dataset (FANToM). We investigate such a discrepancy in \S\ref{sec:analysis_ablation} and \S\ref{sec:analysis_scaling}. 

\section{Analysis}
\label{sec:analysis}

\subsection{High-Order ToM Reasoning}
\label{sec:analysis_high_order}

\begin{figure} [ht!]
    \centering
    \includegraphics[width=\columnwidth]{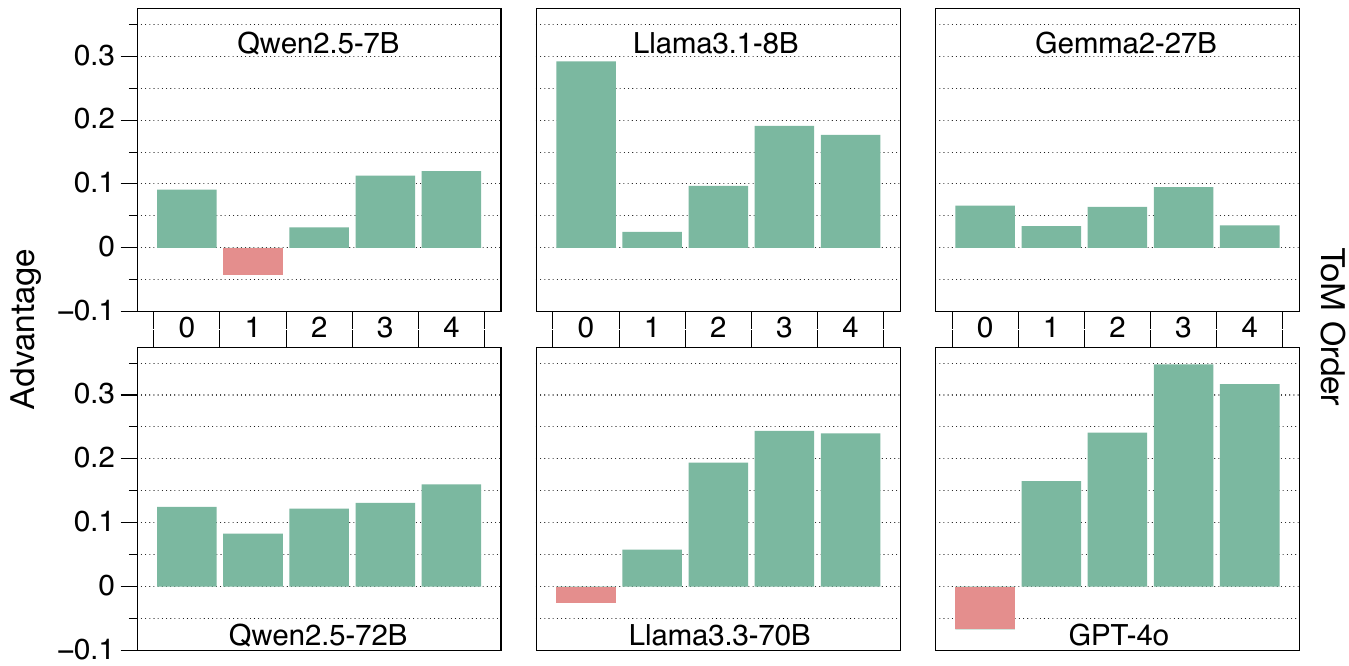}
    \caption{Relative advantage of \ours~on HiToM dataset with respect to ToM order.}
    \label{fig:high_order_hitom}
    \vspace{-0.5em}
\end{figure}

To assess the effectiveness of \ours~in high order ToM reasoning, we analyze its performance on the HiToM dataset, which consists of ToM questions requiring reasoning up to the fourth order. We compute the relative advantage of \ours~with \nkbp~over the zero-shot vanilla prompting baseline. From Figure~\ref{fig:high_order_hitom}, we observe that \ours~ improves mean accuracy across all orders of ToM reasoning, with notable effectiveness in higher-order ToM reasoning. Specifically, results from the Qwen2.5 and Llama3 families demonstrate that \ours~has an increasing advantage as the order of ToM reasoning increases. For the third- and fourth-order ToM reasoning, \ours~achieves an average improvement of $0.160_{\pm0.003}$ and $0.148_{\pm0.004}$ respectively, across all models compared to the baseline. We observe similar trends on ToMi and FANToM albeit they only contain ToM questions up to the second order. See Appendix~\ref{app:high_order_tom} for complete results and analysis on all three datasets.

\subsection{Ablation Study}
\label{sec:analysis_ablation}

To understand the effectiveness of each component of \ours, we conduct an ablation study by (1) keeping the injected knowledge but removing the masking-based perspective-taking mechanism (directly using $\mathcal{E}^{\texttt{aug}}$ as context); and (2) conducting perspective-taking without knowledge injection (applying Equation~\ref{eq:mask} with $G_{\mathcal{E}}$ instead of $G_{\mathcal{E}^{\texttt{aug}}}$).

\paragraph{\nkb~for Perspective Taking}
As shown in Table~\ref{tab:ablation_ki_and_im}, removing the \texttt{IM} mechanism results in an average accuracy drop of $-0.165$ on ToMi, $-0.172$ on HiToM, and $-0.103$ on FANToM. This suggests that the iterative masking mechanism is effective in perspective-taking and crucial for \ours~to achieve boosted performance in ToM reasoning (See Table~\ref{app:tab:ablation_results} for complete results).

\begin{table} [t!]
    \centering 
    \resizebox{\columnwidth}{!}{
        \begin{tabular}{P{0.01\columnwidth} c c c c}
            \toprule
            & & \textbf{\texttt{Llama3.3-70B$^{\texttt{4bit}}$}} & \textbf{\texttt{Qwen2.5-72B$^{\texttt{4bit}}$}} & \textbf{\texttt{GPT-4o}} \\

            \midrule
            \addlinespace[0.5ex]

            \multirow{5}{*}{\rotatebox[origin=c]{90}{ToMi}} & 
            \nkbp & $0.828_{\pm0.012}$ & $0.839_{\pm0.014}$ & $0.847_{\pm0.030}$\\
            & \nkbt & $0.787_{\pm0.024}$ & $0.837_{\pm0.024}$ & $0.795_{\pm0.036}$ \\
            &$\text{w/o }\texttt{KI}$ & \tabbetter{$0.834_{\pm0.067}$} & \tabbetter{$0.845_{\pm0.026}$} & \tabworse{$0.811_{\pm0.028}$} \\
            &$\text{w/o }\texttt{IM}$ & \tabworse{$0.693_{\pm0.014}$} & \tabworse{$0.655_{\pm0.039}$} & \tabworse{$0.674_{\pm0.002}$} \\
            & \textcolor{purewhite}{Eni}$\text{w/o }\texttt{KI,IM}$ & \tabworse{$0.767_{\pm0.015}$} & \tabworse{$0.717_{\pm0.034}$}  & \tabworse{$0.767_{\pm0.041}$}  \\

            \midrule
            \addlinespace[0.5ex]

            \multirow{5}{*}{\rotatebox[origin=c]{90}{HiToM}} & 
            \nkbp & $0.696_{\pm0.007}$ & $0.605_{\pm0.007}$ & $0.733_{\pm0.017}$ \\
            & \nkbt & $0.518_{\pm0.011}$ & $0.473_{\pm0.010}$ & $0.626_{\pm0.020}$ \\
            &$\text{w/o }\texttt{KI}$ & \tabbetter{$0.726_{\pm0.004}$} & \tabbetter{$0.632_{\pm0.003}$} & \tabbetter{$0.751_{\pm0.004}$} \\
            &$\text{w/o }\texttt{IM}$ & \tabworse{$0.460_{\pm0.013}$} & \tabworse{$0.423_{\pm0.008}$} & \tabworse{$0.442_{\pm0.006}$} \\
            & \textcolor{purewhite}{Eni}$\text{w/o }\texttt{KI,IM}$ & \tabworse{$0.534_{\pm0.008}$} & \tabworse{$0.456_{\pm0.012}$} & \tabworse{$0.521_{\pm0.006}$} \\

            \midrule
            \addlinespace[0.5ex]

            \multirow{5}{*}{\rotatebox[origin=c]{90}{FANToM}} & 
            \nkbp & $0.515_{\pm0.020}$ & $0.450_{\pm0.013}$ & $0.531_{\pm0.015}$ \\
            & \nkbt & $0.610_{\pm0.021}$ & $0.574_{\pm0.031}$ & $0.553_{\pm0.011}$ \\
            &$\text{w/o }\texttt{KI}$ & \tabworse{$0.607_{\pm0.018}$} & \tabworse{$0.542_{\pm0.036}$} & \tabworse{$0.539_{\pm0.012}$} \\
            &$\text{w/o }\texttt{IM}$ & \tabworse{$0.500_{\pm0.021}$} & \tabworse{$0.477_{\pm0.017}$} & \tabworse{$0.470_{\pm0.013}$} \\
            & \textcolor{purewhite}{Eni}$\text{w/o }\texttt{KI,IM}$ & \tabworse{$0.486_{\pm0.002}$} & \tabworse{$0.532_{\pm0.025}$} & \tabworse{$0.476_{\pm0.020}$} \\
            \bottomrule
        \end{tabular}
    }
    \caption{Ablation study of \ours~on ToMi, HiToM, and FANToM datasets. "w/o \texttt{KI}" indicates without \textit{entity state knowledge injection}. "w/o \texttt{IM}" denotes without \textit{perspective-taking via iterative masking}. $\improved$ \textit{Improved} and $\worse$ \textit{decreased} results are highlighted.}
    \label{tab:ablation_ki_and_im}
    \vspace{-0.5em}
\end{table}

\paragraph{\nkb~for Knowledge Injection} 
Compared to \texttt{IM} for perspective-taking, entity state knowledge injection is less critical. On ToMi and HiToM, its removal slightly reduces Gemma2-27B’s performance on ToMi but improves performance for all other LLMs on both benchmarks, further highlighting \texttt{IM}’s effectiveness in perspective-taking.
However, for FANToM, entity state knowledge is indispensable, as excluding it results in performance drops across all LLMs. 
For ToMi and HiToM, we hypothesize that larger LLMs are better at handling reporting bias. This aligns with the results shown in Table~\ref{tab:main_results}, where \nkbp~surpasses \nkbt~as LLM size increases, meaning that the fine-grained information about the state of the entity and its causal relationships with events are encapsulated more effectively in the larger LLMs. In such cases, potential inaccuracies in injected entity-state knowledge outweigh its benefits in addressing reporting bias, leading to decreased performance. 
In the case of FANToM, the dialogue-based nature of the dataset makes useful information sparser than in event-based datasets. Here, knowledge injection serves a different role: rather than primarily addressing reporting bias, it compresses important information from utterances into entity-state representation, effectively reducing LLMs' workload in identifying crucial information. See Appendix~\ref{app:ki_examples} for examples.

\begin{table} []
    \centering 
    \resizebox{\columnwidth}{!}{
        \begin{tabular}{c c c c c c}
            \toprule
            Dataset & Model & Precision & Recall & F1-Score \\
            \midrule
            \addlinespace[0.5ex]
            TMi & \texttt{Llama3.3-70B}$^{\texttt{4bit}}$ & 0.859 & 0.968 & 0.910 \\
            \midrule
            \addlinespace[0.5ex]
            FTM & \texttt{Llama3.3-70B}$^{\texttt{4bit}}$ & 0.880 & 0.970 & 0.923 \\
            \bottomrule
        \end{tabular}
    }
    \caption{Performance analysis of key entity recognition in ToMi (TMi) and FANToM (FTM) datasets using \texttt{Llama3.3-70B}$^{\texttt{4bit}}$. See Appendix~\ref{app:annotation} for detailed description of the evaluation process.}
    \label{tab:llm_entity_eval}
\end{table}

\begin{table} []
    \centering 
    \resizebox{\columnwidth}{!}{
        \begin{tabular}{P{0.01\columnwidth} c c c c c}
            \toprule
            & Model & Relevance & Accuracy & Avg. \#Token \\
            \midrule
            \addlinespace[0.5ex]
            \multirow{2}{*}{\rotatebox[origin=c]{90}{TMi}}
            & \nkbt$_\texttt{8B}$ & 0.847 & 0.807 & 7.665 \\
            & \nkbt$_\texttt{70B}$ & 0.870 & 0.860 & 9.740 \\
            \midrule
            \addlinespace[0.5ex]
            \multirow{2}{*}{\rotatebox[origin=c]{90}{FTM}}
            & \nkbt$_\texttt{8B}$ & 0.880 & 0.773 & 8.973 \\
            & \nkbt$_\texttt{70B}$ & 0.880 & 0.700 & 30.517 \\
            \bottomrule
        \end{tabular}
    }
    \caption{Performance analysis of \nkbt~on ToMi (TMi) and FANToM (FTM) datasets. See Appendix~\ref{app:annotation} for detailed description of the evaluation process.}
    \label{tab:nkbt_annotation}
    \vspace{-0.75em}
\end{table}

\begin{table*} [ht!]
    \centering 
    \small
    \resizebox{\linewidth}{!}{
    \begin{tabular} {P{0.01\columnwidth} c c c c c c c c c}
        \toprule
        \addlinespace[0.25ex]
        & & \textbf{\texttt{Qwen2.5-7B}} & \textbf{\texttt{Llama3.1-8B}} & \textbf{\texttt{Gemma2-9B}} & \textbf{\texttt{Gemma2-27B}} & \textbf{\texttt{Llama3.3-70B}$^{\texttt{4bit}}$} & \textbf{\texttt{Qwen2.5-72B}$^{\texttt{4bit}}$} & \textbf{\texttt{GPT-4o}} \\
        \midrule
        \addlinespace[0.5ex]
        \multirow{3}{*}{\rotatebox[origin=c]{90}{ToMi}} & 
        \nkbp & $0.706_{\pm0.044}$ & $0.738_{\pm0.056}$ & $0.865_{\pm0.031}$ & $0.833_{\pm0.018}$ & $0.828_{\pm0.012}$ & $0.839_{\pm0.014}$ & $0.847_{\pm0.030}$ \\
        & \nkbt$_{\texttt{8B}}$ & $0.825_{\pm0.030}$ & $0.796_{\pm0.023}$ & $0.814_{\pm0.020}$ & $0.804_{\pm0.050}$ & $0.787_{\pm0.024}$ & $0.837_{\pm0.024}$ & $0.795_{\pm0.036}$ \\
        & \nkbt$_{\texttt{70B}}$ & \tabbetter{$0.837_{\pm0.017}$} & \tabbetter{$0.835_{\pm0.026}$} & \tabbetter{$0.847_{\pm0.008}$} & \tabbetter{$0.854_{\pm0.023}$} & \tabbetter{$0.844_{\pm0.018}$} & \tabbetter{$0.860_{\pm0.015}$} & \tabbetter{$0.872_{\pm0.026}$} \\

        \midrule
        \addlinespace[0.5ex]

        \multirow{3}{*}{\rotatebox[origin=c]{90}{HiToM}} & 
        \nkbp 
        & 
        $0.508_{\pm0.012}$ & $0.477_{\pm0.005}$ & $0.555_{\pm0.010}$ & $0.576_{\pm0.004}$ & $0.696_{\pm0.007}$ & $0.605_{\pm0.007}$ & $0.733_{\pm0.017}$ \\
        & 
        \nkbt$_{\texttt{8B}}$ &
        $0.457_{\pm0.005}$  & $0.431_{\pm0.010}$ & $0.446_{\pm0.008}$ & $0.478_{\pm0.004}$ & $0.518_{\pm0.011}$ & $0.473_{\pm0.010}$ & $0.626_{\pm0.020}$ \\
        & 
        \nkbt$_{\texttt{70B}}$ & 
        \tabsame{$0.456_{\pm0.007}$} & \tabbetter{$0.444_{\pm0.012}$} & \tabbetter{$0.481_{\pm0.006}$} & \tabbetter{$0.489_{\pm0.012}$} & \tabsame{$0.517_{\pm0.010}$} & \tabworse{$0.467_{\pm0.009}$} & \tabsame{$0.733_{\pm0.010}$} \\

        \midrule
        \addlinespace[0.5ex]

        \multirow{3}{*}{\rotatebox[origin=c]{90}{FANToM}} & 
        \nkbp & $0.445_{\pm0.026}$ & $0.442_{\pm0.018}$ & $0.439_{\pm0.023}$ & $0.462_{\pm0.014}$ & $0.515_{\pm0.020}$ & $0.450_{\pm0.013}$ & $0.531_{\pm0.015}$\\
        & \nkbt$_{\texttt{8B}}$ & $0.487_{\pm0.018}$ & $0.545_{\pm0.036}$ & $0.530_{\pm0.012}$ & $0.582_{\pm0.028}$ & $0.610_{\pm0.021}$ & $0.574_{\pm0.031}$ & $0.553_{\pm0.011}$ \\
        & \nkbt$_{\texttt{70B}}$ & \tabworse{$0.479_{\pm0.032}$} & \tabworse{$0.517_{\pm0.041}$} & \tabworse{$0.491_{\pm0.039}$} & \tabworse{$0.529_{\pm0.028}$} & \tabworse{$0.537_{\pm0.011}$} & \tabworse{$0.494_{\pm0.023}$} & \tabworse{$0.539_{\pm0.012}$}\\
        \bottomrule
    \end{tabular}
    }
    \caption{Results of scaling \nkbt~from Llama3.1-8B to Llama3.3-70B. $\improved$ \textit{Improved}, $\unchanged$ \textit{Unchanged}, and $\worse$ \textit{decreased} results are highlighted in the corresponding color.}
    \label{tab:ablation_nkbt}
    \vspace{-1em}
\end{table*}

\subsection{Effectiveness of LLMs and \nkb}
\label{sec:analysis_scaling}
\begin{figure} [ht!]
    \centering
    \includegraphics[width=\columnwidth]{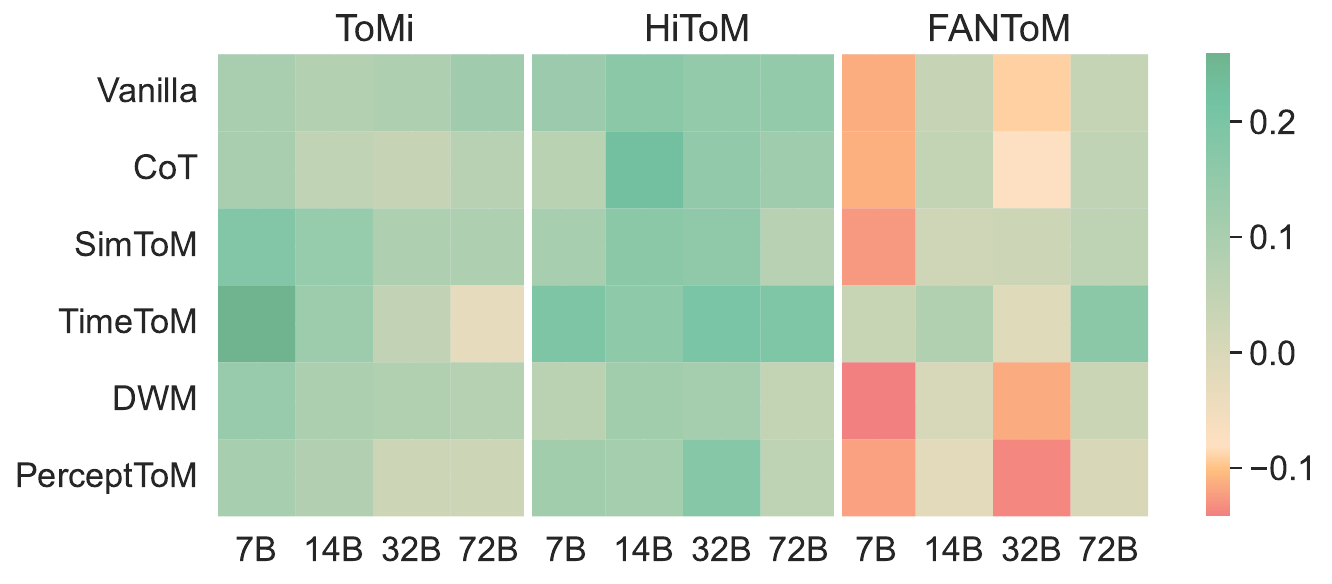}
    \caption{Relative advantage of \ours~on ToMi, HiToM, and FANToM datasets. We use \nkbp~as the pivot method for ToMi and HiToM and \nkbt~for FANToM. Exact mean accuracies are shown in Table~\ref{app:tab:scaling_results}. Model sizes shown in x-axis are Qwen2.5 models.}
    \label{fig:scaling_heatmap}
    \vspace{-0.75em}
\end{figure}

The effectiveness of \ours~is contingent upon the capabilities of both the \nkb~neural knowledge base and the LLM deployed in the framework. In this section, we conduct analysis of \ours~with an aim to explore the following two questions: (1) Does \ours~benefit LLMs of larger size? and (2) How effective is our \nkb neural knowledge base and does scaling \nkb lead to increased performance? 

We raise the first question by hypothesizing that perspective-taking, albeit the challenges posed by its multi-hop nature, could become solvable by more capable LLMs. Empirically speaking, the capability of LLMs positively correlates to their number of parameters. To eliminate potential confounding factors, we analyze the effectiveness of LLMs from the Qwen2.5 family with sizes ranging from 7B to 72B \cite{qwen2.5}. 

In the second question, we aim to examine the effectiveness of \nkbt. Trained using data from OpenPI2.0, we wish to investigate how well can the knowledge encapsulated in \nkbt~be transferred to aid ToM reasoning. Aside from the performance of \nkbt, we also explore the effectiveness of scaling of \nkbt. In addition to the \nkbt~used in previous experiments, which is trained using a Llama3.1-8B model, we trained another \nkbt~based on Llama3.3-70B, which we denote as \nkbt$_\texttt{70B}$. Experiments with \nkbt$_\texttt{70B}$ are carried out following the same procedure described in \S\ref{sec:exp}.

\paragraph{Scaling of Base LLMs} We compute the relative advantage of \ours~by calculating the difference in mean accuracy between \ours~and the most performant baseline methods (see Table~\ref{tab:main_results}). We use \nkbp~as the pivot method for ToMi and HiToM and \nkbt~for FANToM. Figure~\ref{fig:scaling_heatmap} shows two trends: (1) a slight diminishment in advantage on ToMi and (2) a gradual increase in advantage on FANToM. We attribute this to the differing difficulty levels of these two datasets. ToMi, which consists of short sequences of concise events, becomes easier to solve with large-scale LLMs. Conversely, FANToM, featuring long sequences of lengthy dialogues, remains challenging even for larger LLMs. HiToM, positioned between these two extremes with long sequences of concise events, shows that \ours~has a consistent advantage regardless of the LLM sizes. This discrepancy in performance and model scaling effect between ToMi and FANToM aligns with the analysis in \S\ref{sec:exp} and \S\ref{sec:analysis_scaling}. These findings suggest that while prompting large-scale LLMs can potentially tackle ToM reasoning involving short event sequences (as in ToMi), ToM reasoning about lengthy event or dialogue sequences (as in HiToM and FANToM) can benefit from the fine-grained entity state knowledge as well as the symbolic masking mechanism of \ours.

\paragraph{Effectiveness of LLMs in Recognizing Key Entity-Attributes}
Recognizing entities and their attributes (Equation~\ref{eq:eoi_query}) that are indispensable for answering the ToM questions posed is a critical pre-requisite for the effectiveness of \ours. On the one hand, failing to recognize a key entity will disable \ours~to properly augment the events with critical information. On the other hand, erroneously identify an extraneous entity will lead to inclusion of redundant information, which will prolong the context and increase the reasoning burden of the LLM. To evaluate the quality of key entities and attributes extracted by LLMs, we manually labeled 300 entities and attributes identified using \texttt{Llama3.3-70B}$^{\texttt{4bit}}$. Evaluation results from Table~\ref{tab:llm_entity_eval} suggests that LLMs are more than competent in identifying key entities and attributes. With a F1-score of 0.910 on the ToMi dataset and 0.923 on the FANToM dataset, it is safe to conclude that the vast majority of entities identified by LLMs are indeed vital to answering the ToM questions posed. 

\paragraph{Effectiveness of \nkb~in Generating Entity State Information}
To understand the effectiveness of scaling \nkb, we trained two \nkbt~models using the same OpenPI2.0 dataset. With Llama3.1-8B as the base model, we trained \nkbt$_{\texttt{8B}}$. Further, with Llama3.3-70B, we trained \nkbt$_{\texttt{70B}}$\footnote{Training details are provided in Appendix~\ref{app:training_nkbt}.}. Table~\ref{tab:ablation_nkbt} shows that there is an obvious discrepancy between the scaling effect of \nkbt: \nkbt$_{\texttt{70B}}$ consistently outperforms \nkbt$_{\texttt{8B}}$ in ToMi and HiToM while underperforming \nkbt$_{\texttt{8B}}$ in FANToM.  
\begin{itemize}[leftmargin=*, noitemsep]
    \item \textbf{Relevance}: whether the entity and attribute contribute to answering the ToM question. This evaluates the same aspects of \ours~as the precision scores shown in Table~\ref{tab:llm_entity_eval}
    \item \textbf{Accuracy}: whether the entity state can be inferred from the given context. 
\end{itemize}
From Table~\ref{tab:nkbt_annotation}, we see that the relevance scores of both ToMi and FANToM exceed 80\%, indicating that Llama3.3-70B is capable of identifying entities and attributes useful for ToM reasoning. Entity state length and accuracy are closely correlated. \nkbt$_\texttt{70B}$ produces a more articulated response compared to its counterpart. Specifically, scaling \nkbt~brings 5.3\% improvement in accuracy on ToMi while increasing response length by only 2.075 tokens. In contrast, FANToM experiences a significant 21.544 token increase in response length, which reduces \nkbt's efficiency as an information compressor and leads to greater hallucination, resulting in a 7.3\% drop in accuracy. We provide demonstration examples in Appendix~\ref{app:nkbt_comparisons}.

\section{Conclusion}
In this work, we introduced \ours, a neuro-symbolic framework designed to enhance the ToM reasoning capabilities of LLMs. By leveraging a Neural Knowledge Base of Entity States through an iterative masking mechanism and knowledge injection, \ours~accomplishes the bulk of ToM reasoning via perspective-taking through symbolic reasoning, which alleviates LLMs' reasoning burden. Experimental results across multiple benchmarks demonstrate that \ours~outperforms existing methods, particularly excelling in high-order ToM reasoning scenarios.
Our analysis highlights the effectiveness of the iterative masking mechanism in maintaining strong performance across varying depths of ToM reasoning, as well as the critical role of fine-grained entity state knowledge in compressing key information in complex event sequences (as in FANToM). Furthermore, the framework's efficiency and scalability make it a promising solution for addressing the computational challenges associated with high-order ToM reasoning tasks.

\section*{Limitations}

\paragraph{ToM Reasoning Beyond Character Perception}
\ours~tackles ToM reasoning of characters' beliefs based on their perceptions. While we believe that reasoning about characters' perceptions serve as a cornerstone for all types of ToM reasoning, future work may explore methods to facilitate real-world ToM reasoning about characters' emotions, intentions, desires, and their inherent subjectivity \cite{zhou2025modeling}. 

\paragraph{Neural Knowledge Base}
 \ours~relies on access to a Neural Knowledge Base (NKB) to retrieve entity-state information for answering ToM questions. While Table~\ref{tab:nkbt_annotation} shows that \nkb~is capable of producing accurate entity-state information, it can be further improved (e.g. full-parameter fine-tuning instead of LoRA). Further, expanding the NKB to incorporate richer entity-state details, including emotional, temporal, and causal relationships, would be beneficial for ToM reasoning about high-level information.

 \paragraph{Error Propogation}
 While experiments demonstrate the effectiveness of the \texttt{IM} mechanism, it is prone to error propagation. In the case of high-order ToM reasoning, applying a wrong mask in the iterative masking process will lead to the event being erroneously excluded and vice versa. Additionally, in cases requiring complex reasoning about non-linear or intertwined event dependencies, the symbolic Iterative Masking (IM) mechanism may need to be enhanced. 

 \section*{Ethics Statement}
This study aims to enhance LLMs' ToM reasoning by improving the accuracy and efficiency of perceptual perspective-taking, ultimately optimizing their effectiveness in communication. ToM reasoning is essential for enhancing LLMs' ability to interact with humans (e.g., in chatbots) or other LLMs (e.g., in multi-agent systems). The evaluation datasets used in this study have been peer-reviewed and widely adopted in previous research. However, these datasets may introduce issues such as cultural bias and often lack demographic information. Future research could incorporate auxiliary data, such as demographic and personality traits, to improve representativeness across diverse ethnic and cultural backgrounds.

\section*{Acknowledgments}
This work was supported in part by the UK Engineering and Physical Sciences Research Council (EPSRC) through an iCASE award with Huawei London Research Centre and a Turing AI Fellowship (grant no. EP/V020579/1, EP/V020579/2).

\bibliography{custom}

\clearpage
\appendix
\setcounter{table}{0}
\renewcommand{\thetable}{A\arabic{table}}
\setcounter{figure}{0}
\renewcommand{\thefigure}{A\arabic{figure}}

\section{Data Examples}
\label{app:data_examples}
See below for examples of event sequences/dialogues and ToM questions from ToMi (Box~\ref{tomi_examples}), HiToM (Box~\ref{hitom_examples}), and FANToM (Box~\ref{fantom_examples}) datasets.

\begin{tcolorbox}[
    enhanced,
    attach boxed title to top center={
        yshift=-3mm,yshifttext=-1mm
    },
    breakable,
    colframe=yellow!75!black,
    colbacktitle=yellow,
    title=ToMi Example,
    coltitle=black,
    collower=purple, 
    label=tomi_examples,
    fonttitle=\bfseries,
    boxed title style={size=small,colframe=yellow!50!black}
    ]
    Benjamin entered the crawlspace. \\
    Abigail entered the crawlspace. \\
    Emily entered the crawlspace. \\
    The t-shirt is in the cupboard. \\
    The cupboard is in the crawlspace. \\
    Abigail exited the crawlspace. \\
    Emily moved the t-shirt to the basket. \\
    The basket is in the crawlspace. \\
    Benjamin hates the coat. \\
    Emily exited the crawlspace. \\
    Abigail entered the crawlspace.
    \tcblower
    <$0^{th}$-Order (Factual) Question>  \\
    Where is the t-shirt in the begining? \\
    \\
    <$1^{st}$-Order ToM Question> \\
    Where will Abigail search for the t-shirt? \\
    \\
    <$2^{nd}$-Order ToM Question> \\
    Where does Benjamin think that Abigail search for the t-shirt? 
\end{tcolorbox}

\begin{tcolorbox}[
    enhanced,
    attach boxed title to top center={
        yshift=-3mm,yshifttext=-1mm
    },
    breakable,
    colframe=yellow!75!black,
    colbacktitle=yellow,
    title=HiToM Example,
    coltitle=black,
    collower=purple, 
    label=hitom_examples,
    fonttitle=\bfseries,
    boxed title style={size=small,colframe=yellow!50!black}
    ]
    William, Lily, Aiden, Emma and Isla entered the porch. \\
    The melon is in the green bathtub. \\
    William made no movements and stayed in the porch for 1 minute. \\
    William exited the porch. \\
    Lily made no movements and stayed in the porch for 1 minute. \\
    Lily exited the porch. \\
    Aiden moved the melon to the blue pantry. \\
    Lily likes the green bucket. \\
    Aiden exited the porch. \\
    Emma moved the melon to the green bucket. \\
    Emma exited the porch. \\
    Isla moved the melon to the green bathtub. \\
    Isla exited the porch. \\
    William, Lily, Aiden, Emma and Isla entered the waiting room.
    Emma, Isla and Aiden entered the basement. \\
    The watermelon is in the blue suitcase. \\
    Emma made no movements and stayed in the basement for 1 minute.
    Emma exited the basement. \\
    Isla moved the watermelon to the blue bottle. \\
    Isla exited the basement. \\
    Aiden made no movements and stayed in the basement for 1 minute. \\
    Aiden exited the basement. \\
    Emma, Isla and Aiden entered the waiting room. \\
    Aiden likes the red pantry. \\
    William, Isla, Aiden and Lily entered the front yard. \\
    The beans is in the red bottle. \\
    William made no movements and stayed in the front yard for 1 minute. \\
    Isla likes the green cupboard. \\
    William exited the front yard. \\
    Isla moved the beans to the green bottle. \\
    Isla exited the front yard. \\
    Aiden made no movements and stayed in the front yard for 1 minute. \\
    Aiden exited the front yard. \\
    Lily made no movements and stayed in the front yard for 1 minute. \\
    Lily exited the front yard. \\
    William, Isla, Aiden and Lily entered the waiting room.
    \tcblower
    <$0^{th}$-Order (Factual) Question>  \\
    Where is the melon really? \\
    \\
    <$1^{st}$-Order ToM Question> \\
    Where does William really think the melon is? \\
    \\
    <$2^{nd}$-Order ToM Question> \\
    Where does Lily think William thinks the melon is? \\
    \\
    <$3^{rd}$-Order ToM Question> \\
    Where does Emma think Lily thinks William thinks the melon is? \\
    \\
    <$4^{th}$-Order ToM Question> \\
    Where does Isla think Emaa thinks Lily thinks William thinks the melon is?
\end{tcolorbox}

\begin{tcolorbox}[
    enhanced,
    attach boxed title to top center={
        yshift=-3mm,yshifttext=-1mm
    },
    breakable,
    colframe=yellow!75!black,
    colbacktitle=yellow,
    title=FANToM Example,
    coltitle=black,
    collower=purple, 
    label=fantom_examples,
    fonttitle=\bfseries,
    boxed title style={size=small,colframe=yellow!50!black}
    ]
    Armani: Hi Troy, it's nice to meet you. What's been your experience in maintaining good mental health while in a relationship? \\

    Troy: Hey Armani! I've found that the most important thing for me is understanding that I need to take care of my own mental health first. I look at it like the whole oxygen mask in an airplane situation- you have to secure your own before helping someone else. \\

    Armani: That's an interesting perspective, it's all about maintaining individual wellness before being able to fully contribute to a relationship. I've always subscribed to the idea of communication being an integral part of it too. Being open about my mental health issues with my partner has always helped to build understanding. \\

    Troy: I definitely see the merit in that too. It can be hard to open up about these things sometimes, especially if the other person doesn't fully understand. \\

    Armani: Absolutely, I think it's important for both partners to constantly educate themselves on each other's mental health issues. It not only encourages empathy but also helps in mitigating unnecessary tensions. \\

    Troy: You're right. From my experience, I've found that maintaining a healthy work-life balance is also essential. Stress from work can really take a toll on my mental health, and it's hard to keep that stress from affecting my relationships. \\

    Armani: I completely agree, Troy. Ignoring the effects of work-related stress on our mental wellbeing can have dire consequences on our relationships. Just as we wouldn't like to bring home a flu virus, we shouldn't be infecting our home with stress either. It's all about creating boundaries. \\

    Troy: Absolutely. It's great to find someone else who understands the importance of maintaining mental health while in a relationship. It's definitely a balance, but it's worth it in the long run. \\

    Armani: Couldn't agree more Troy. I'm glad we could have this open conversation about such an important topic. The more we talk, the more we can break the stigma surrounding mental health issues. \\

    Cynthia: Hello Troy, Armani. The two of you have been engaged in quite a meaningful conversation, it seems. \\

    Armani: Hi Cynthia, yes indeed. We've been discussing the importance of good mental health maintenance in a relationship. \\

    Cynthia: Such a crucial topic! In my experience, clear actionable boundaries have played a very big role in mental wellness. Making sure me and my partner are on the same page about our needs and wants can genuinely de-stress the environment. \\

    Troy: Couldn't agree more, Cynthia. The clear establishment of boundaries helps a lot in maintaining a harmonious balance. \\

    Armani: Definitely, Cynthia. It's such a simple concept yet so often overlooked. People sometimes shy away from setting boundaries, afraid it might upset the other person. But it's needed for mutual respect and understanding. \\

    Cynthia: Yes, Armani. And it's these boundaries that create stronger communication channels. So, when I'm experiencing a difficult time with my mental health, I find it easier to express to my partner. \\

    Troy: That really hits home, Cynthia. It's incredibly liberating to be able to express ourselves without fear of judgement. \\

    Armani: Absolutely, Troy! And the thing is, this whole conversation really highlights the importance of communication. Everything, from understanding personal mental health, setting boundaries, to dealing with stress, involves communicating effectively. \\

    Cynthia: Here's to hoping that more people can learn and implement these practices in their relationships. Good mental health is so important, and talking about it openly like this is a great step in the right direction. 
    \tcblower
    <$0^{th}$-Order (Factual) Question>  \\
    What did Armani and Troy discuss as a preventative measure against work-related stress affecting their re       lationships? \\
    \\
    <$1^{st}$-Order ToM Question> \\
    What does Cynthia believe are the necessary steps suggested by Armani and Troy for dealing with mental health issues in a relationship? \\
    \\
    <$2^{nd}$-Order ToM Question> \\
    What does Armani believe about Cynthia's belief regarding the necessary steps suggested by Armani and Troy for dealing with mental health issues in a relationship? \\
    Where does Benjamin think that Abigail search for the t-shirt? 
\end{tcolorbox}

\section{Training of \nkbt}
\label{app:training_nkbt}

We train \nkbt~based on Llama3.1-8B and Llama3.3-70B using OpenPI2.0 dataset. The training of Llama3.1-8B is done using Low-Rank Adaptation \cite{hu2022lora} and the training of Llama3.3-70B is done using Quantized Low-Rank Adaptation \cite{dettmers2024qlora}. The rank of the decomposed matrix is set to $\texttt{LoRA-Rank}=32$. We use a batch size of 64 and a gradient accumulation of 2 steps, leading to an effective batch size of 128. The learning rate is initially set to 5e-5 and adjusted using a Cosine Annealing with a warm-up ratio of 0.01. Each model is trained for 3 epoches. The training is done on 2 NVIDIA A100$^{80\texttt{GB}}$ GPUs using LlamaFactory \cite{llamafactory}.

We modified the formatting convention used in OpenPI \cite{tandon-etal-2020-dataset}. In the original formatting, the output is formatted as 
\[
    \begin{aligned}
    [\texttt{Attribute}]~\text{of}~[\texttt{Entity}]~\text{is}~[\texttt{Previous State}] \\
    \text{before and}~[\texttt{Current State}]~\text{afterwards}.
    \end{aligned}
\]
In this formulation, the model is tasked to predict both the \texttt{previous state} as well as the \texttt{current state} after the event has taken place. We modify this formulation by removing the \texttt{previous state} and only asking the model to predict the \texttt{current state}:
\[
    [\texttt{Attribute}]~\text{of}~[\texttt{Entity}]~\text{becomes}~[\texttt{State}]
\]
this modification is made to alleviate model's reasoning burden and to focus on deriving the effect the the event. 

The events are provided to the model in a cumulative fashion. For instance, when generating entity state knowledge for the $i$-th event, the model is provided with the events from the first event to the $i$-th event.

\section{Transforming High-order ToM Question to First-order}
\label{app:reduce_tom_question_order}

As discussed in \citet{hou-etal-2024-timetom}, perspective-taking reduces the reasoning depth of high-order ToM questions. Therefore, we can transform high-order ToM questions into first-order questions after perspective-taking. We first justify how \ours~enables reduction in ToM reasoning order and then provide a description of the question transformation process.

\subsection{ToM Order Reduction}
ToM order reduction is possible thanks to the \texttt{IM} mechanism, which carries out the high-order, multi-hop ToM reasoning process symbolically using spatial scene graphs. We use an example to elicit the necessity of \texttt{IM} mechanism. 

Consider the following event sequence:
\vspace{0.5em}
\begin{mdframed}[style=MyQuoteFrame]
\begin{itemize}[leftmargin=6mm, noitemsep]
\item William, Lily, Aiden, Emma and Isla entered the porch.
\item The melon is in the green bathtub.
\item Aiden moved the melon to the blue pantry.
\item Lily likes the green bucket.
\item Aiden exited the porch.
\item Lily made no movements and stayed in the porch for 1 minute.
\item Lily exited the porch.
\item Emma moved the melon to the green bucket.
\item Emma exited the porch.
\item Isla moved the melon to the green bathtub.
\item Isla exited the porch.
\item William moved the melon to the red bucket.
\item William exited the porch.
\item William, Lily, Aiden, Emma, and Isla entered the waiting room.
\end{itemize}
\end{mdframed}

Consider the following third-order ToM question:

\noindent
\textit{Where does Emma think Lily thinks William thinks the melon is?}

\noindent
\textbf{Answer}: blue pantry

\vspace{1em}

The above question can be reduced to the following first-order ToM question:

\noindent
Where does William think the melon is?

\noindent
\textbf{Answer}: red bucket

\vspace{0.5em}
Notice that the answers to the third-order and first-order ToM question are different. This is because answering the original question requires three reasoning hops:
\begin{itemize}[leftmargin=3.5em, noitemsep]
    \item[\textbf{Step 1}] Infer Emma’s belief about the environment.
    \item[\textbf{Step 2}] Given Emma’s belief, infer Lily’s belief about the environment.
    \item[\textbf{Step 3}] Given Emma’s belief of Lily’s belief, infer William’s belief about the environment.
\end{itemize}

\noindent
In contrast, a first-order ToM question only requires a single reasoning hop:
\begin{itemize}[leftmargin=3.5em, noitemsep]
    \item[\textbf{Step 1}] Infer William’s belief about the environment.
\end{itemize}

To alleviate the multi-hop reasoning burden from LLMs, we utilize the spatial scene graphs with the \texttt{IM} mechanism. Each character’s spatial scene graph can be treated as a reasoning hop. The aggregates of the scene graphs through \texttt{IM} effectively performs multi-hop reasoning symbolically. This allows us to transform a high-order ToM question into a simpler one while preserving accuracy.

For example, after \texttt{IM}, irrelevant events are masked (striked through), leaving only those representing Emma’s belief of Lily’s belief of William’s belief:
\vspace{0.5em}
\begin{mdframed}[style=MyQuoteFrame]
\begin{itemize}[leftmargin=6mm, noitemsep]
\item William, Lily, Aiden, Emma and Isla entered the porch.
\item The melon is in the green bathtub.
\item Aiden moved the melon to the blue pantry.
\item Lily likes the green bucket.
\item Aiden exited the porch.
\item Lily made no movements and stayed in the porch for 1 minute.
\item Lily exited the porch.
\item \st{Emma moved the melon to the green bucket}.
\item \st{Emma exited the porch.}
\item \st{Isla moved the melon to the green bathtub.}
\item \st{Isla exited the porch.}
\item \st{William moved the melon to the red bucket.}
\item \st{William exited the porch.}
\item \st{William, Lily, Aiden, Emma, and Isla entered the waiting room.}
\end{itemize}
\end{mdframed}

By reducing the original third-order ToM question into that of first-order, we are able to obtain the correct answer:

\vspace{0.5em}
\noindent
\textbf{Original Third-order ToM Question} \\
\textit{Where does Emma think Lily thinks William thinks the melon is?} \\
\textbf{Reduced First-order ToM Question} \\
\textit{Where does William think the melon is?} \\
\textbf{Answer:} blue pantry

\subsection{Question Transformation}
We transform the original high-order ToM questions to first-order by prompting LLMs with 5-shot demonstrations for each dataset. The transformed questions are used for QA with \ours~as well as TimeToM. See below for an illustrative example of transforming a third-order ToM question into a first-order question:

\vspace{0.5em}
\noindent
\textbf{3$^{rd}$-Order $\rightarrow$ 1$^{st}$-Order ToM Question:} \\
\textcolor{purewhite}{TEST} \textit{Where does Emma think Lily thinks William thinks the melon is?} \\
\textcolor{purewhite}{TESTTESTTESTTEST}$\Longrightarrow$ \\
\textcolor{purewhite}{TEST} \textit{Where does William think the melon is?}

\section{Prompts}
\label{app:prompt_bank}
The following prompt is used to infer key entities and attributes from an event sequence and ToM question as described in \S\ref{sec:framework_nkb} and Equation~\ref{eq:eoi_query}.
\begin{tcolorbox}[
    enhanced,
    attach boxed title to top center={
        yshift=-3mm,yshifttext=-1mm
    },
    breakable,
    colframe=green1!75!black,
    colbacktitle=green1,
    title=Infer Key Entities and Attributes,
    coltitle=black,
    collower=black, 
    label=tomi_ki_example_part2,
    fonttitle=\bfseries,
    boxed title style={size=small,colframe=yellow!50!black}
    ]
    <Events> \\
    \{\{indexed narrative\}\} \\
  
    <Questions> \\
    \{\{question list\}\} \\
  
    Based on the list of <questions>, extract at most five entities and their attributs that are needed for answering the questions. Note that one entity could corresponds to multiple attributes. List the entities and their attributes. For instance, if the "location of tie", "placement of tie", and "color of crate" are important for answering the questions, the response should be formatted as follows: \\
  
    <entities> \\
    - location of tie \\
    - placement of tie \\
    - color of crate \\
    </entities> \\
  
    You must include at least one entity that is not a person. Only extract entities directly mentioned in the questions, do not make any further inference. Do not include any entities that indicate a time or point in time. First briefly reason about the content of the events and the questions and then provide a comprehensive list of at most five entities and their attributes with the following format: \\
  
    <entities> \\
    - attribute of entity \\
    - attribute of entity \\
    ... \\
    </entities>
\end{tcolorbox}

We use the following prompt to identify locations appeared in the event sequence. The location information is used as anchor when constructing the spatial scene graphs to map the locations produced by \nkb~to a common location space.

\begin{tcolorbox}[
    enhanced,
    attach boxed title to top center={
        yshift=-3mm,yshifttext=-1mm
    },
    breakable,
    colframe=green1!75!black,
    colbacktitle=green1,
    title=Extract Locations from Events,
    coltitle=black,
    collower=black, 
    label=tomi_ki_example_part3,
    fonttitle=\bfseries,
    boxed title style={size=small,colframe=yellow!50!black}
    ]
    <Events>\\
    \{\{indexed narrative\}\}\\
    
    What are the rooms mentioned in these events? List all the rooms in the following format: \\
    - Room1 \\
    - Room2 \\
    ... \\
    
    Please exclude entities in which people cannot enter. Each narrative must contain at least one room and your answer must include at least one room. Provide your answer as bullet points without any explanation.
\end{tcolorbox}

The following prompt is used to acquire the entity state knowledge by prompting LLMs (\nkbp).
\begin{tcolorbox}[
    enhanced,
    attach boxed title to top center={
        yshift=-3mm,yshifttext=-1mm
    },
    breakable,
    colframe=green1!75!black,
    colbacktitle=green1,
    title=Generate Entity State Knowledge,
    coltitle=black,
    collower=black, 
    label=tomi_ki_example_part4,
    fonttitle=\bfseries,
    boxed title style={size=small,colframe=yellow!50!black}
    ]
    <Events>\\
    \{\{indexed narrative\}\} \\
    
    <Entity-of-Interest> \\
    \{\{eoi list\}\} \\ 
    Given the list of events and entities-of-interest, track the state of the attribute of entities throughout the events. Generate state of each attribute of entities as a list in the following format: \\
    - [Event Index]: [Entity Attribute] becomes [State] \\
    - [Event Index]: [Entity Attribute] becomes [State] \\
    ... \\
    
    Determine spatial information according to the following instructions: \\
    - All of the location changes, if exist, are explicitly stated in the events. \\
    - If the event does not state that a character left a room, assume that the character remains in the previous location. \\

    Generate the answers exactly as instructed without any explanation or note. Only generate the event indices where there is a entity state change, omit other events.
\end{tcolorbox}

\section{Visualization of Complexity}
\label{app:vis_complexity}
\begin{figure} [ht!]
    \centering
    \includegraphics[width=\columnwidth]{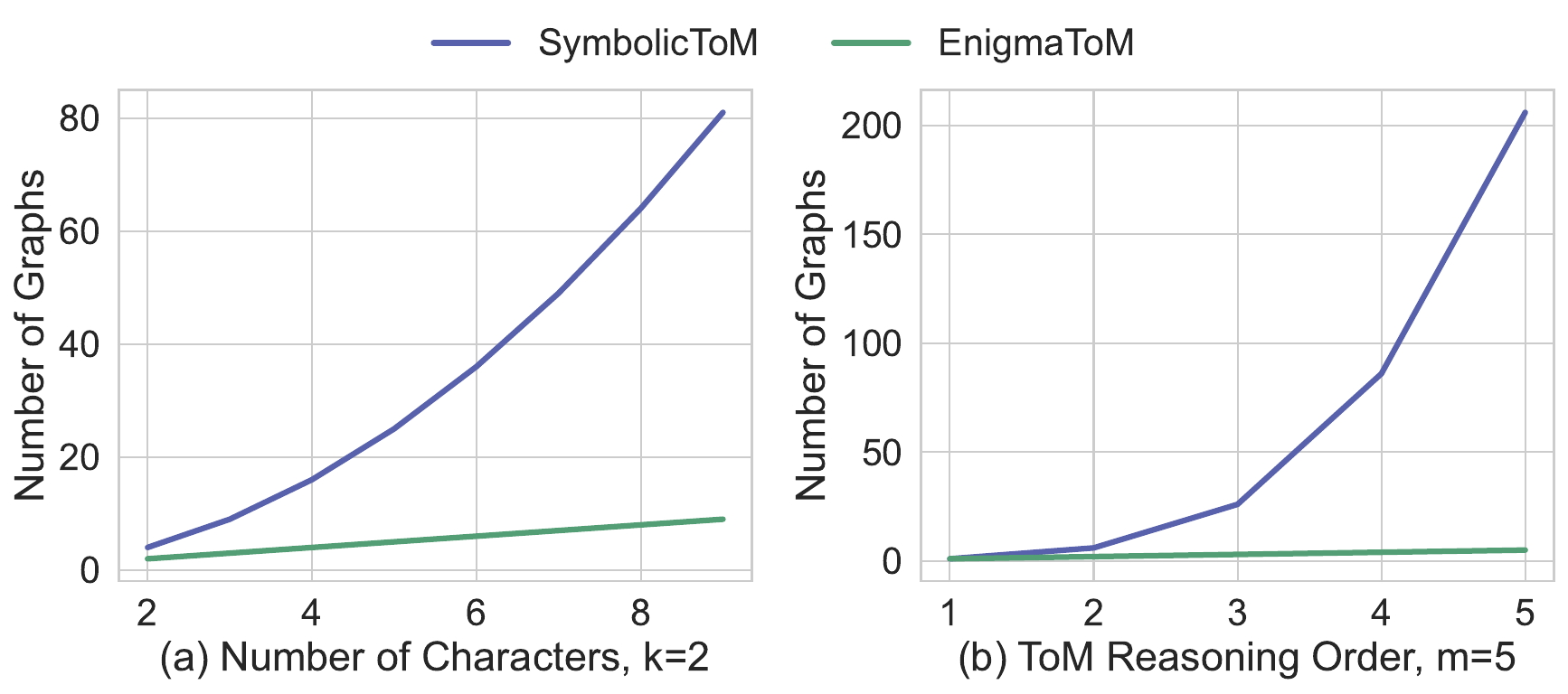}
    \caption{Visualization of the complexity of SymbolicToM and \ours~with respect to the number of belief graphs need to be constructed.}
    \label{fig:complexity_vis}
\end{figure}

In \S\ref{sec:framework_efficiency}, we provide a general analysis of the complexity of SymbolicToM and \ours~with respect to the number of belief graphs need to be constructed. Here, we provide a visualization to better demonstrate the efficiency of the \texttt{IM}-based perspective-taking method of \ours. Figure~\ref{fig:complexity_vis} shows the number of belief graphs need to be constructed for SymbolicToM and \ours~with respect to the number of characters (Part a) or the order of ToM reasoning (Part b). When plotting against the number of character, we fix the ToM reasoning order to be $k=2$, which requires a minimum of 2 characters. When plotting against the Tom reasoning order, we fix the number of characters to be $m=5$, which supports up to $5^{th}$-order acyclic ToM reasoning. 

\begin{table*} [h]
    \centering
    \small
    \resizebox{0.75\linewidth}{!}{
    \begin{tabular} {P{0.001\columnwidth} c c c c c c }
        \toprule
        & & Qwen2.5-7B & Qwen2.5-14B & Qwen2.5-32B & Qwen2.5-72B$^{\text{4bit}}$ \\
        \midrule
        \addlinespace[0.5ex]
        \multirow{8}{*}{\rotatebox[origin=c]{90}{ToMi}} & 
        VNL & $0.722_{\pm0.045}$ & $0.747_{\pm0.053}$ & $0.720_{\pm0.036}$ & $0.717_{\pm0.034}$ \\
        & CoT & \underbar{$0.724_{\pm0.026}$} & \underbar{$0.773_{\pm0.030}$} & $0.769_{\pm0.029}$ & $0.767_{\pm0.033}$ \\
        & SToM & $0.642_{\pm0.022}$ & $0.686_{\pm0.042}$ & $0.721_{\pm0.029}$ & $0.749_{\pm0.020}$ \\
        & TToM & $0.567_{\pm0.024}$ & $0.699_{\pm0.027}$ & $0.758_{\pm0.019}$ & \boldmath{$0.865_{\pm0.018}$} \\
        & DWM  & $0.686_{\pm0.023}$ & $0.732_{\pm0.033}$ & $0.723_{\pm0.053}$ & $0.762_{\pm0.051}$ \\
        & PToM & $0.720_{\pm0.038}$ & $0.743_{\pm0.050}$ & $0.783_{\pm0.021}$ & $0.809_{\pm0.033}$ \\
        \midrule
        \addlinespace[0.5ex]
        \rowcolor{tablewhite} & \textbf{\nkbp} & $0.706_{\pm0.044}$ & $0.736_{\pm0.019}$ & \boldmath{$0.828_{\pm0.009}$} & \underbar{$0.839_{\pm0.014}$} \\
        \rowcolor{tablewhite} & \textbf{\nkbt} & \boldmath{$0.825_{\pm0.030}$} & \boldmath{$0.826_{\pm0.027}$} & \underbar{$0.809_{\pm0.027}$} & $0.837_{\pm0.024}$ \\

        \addlinespace[0.5ex]
        \bottomrule
        \addlinespace[0.5ex]

        \multirow{8}{*}{\rotatebox[origin=c]{90}{HiToM}} & 
        VNL & $0.378_{\pm0.013}$ & $0.389_{\pm0.014}$ & $0.524_{\pm0.007}$ & $0.456_{\pm0.012}$ \\
        & CoT & $0.441_{\pm0.007}$ & $0.330_{\pm0.020}$ & $0.523_{\pm0.009}$ & $0.481_{\pm0.011}$ \\
        & SToM & $0.402_{\pm0.009}$ & $0.389_{\pm0.025}$ & $0.520_{\pm0.007}$ & $0.536_{\pm0.018}$ \\
        & TToM & $0.316_{\pm0.010}$ & $0.397_{\pm0.017}$ & $0.473_{\pm0.011}$ & $0.415_{\pm0.013}$ \\
        & DWM  & $0.444_{\pm0.020}$ & $0.436_{\pm0.019}$ & \underbar{$0.566_{\pm0.011}$} & \underbar{$0.560_{\pm0.009}$} \\
        & PToM  & $0.393_{\pm0.018}$ & \underbar{$0.446_{\pm0.017}$} & $0.500_{\pm0.011}$ & $0.548_{\pm0.016}$ \\
        \midrule
        \addlinespace[0.5ex]
        \rowcolor{tablewhite} & \textbf{\nkbp} & \boldmath{$0.508_{\pm0.012}$} & \boldmath{$0.554_{\pm0.013}$} & \boldmath{$0.674_{\pm0.011}$} & \boldmath{$0.605_{\pm0.007}$} \\
        \rowcolor{tablewhite} & \textbf{\nkbt}  & \underbar{$0.457_{\pm0.005}$} & $0.414_{\pm0.013}$ & $0.504_{\pm0.013}$ & $0.473_{\pm0.010}$ \\

        \addlinespace[0.5ex]
        \bottomrule
        \addlinespace[0.5ex]

        \multirow{8}{*}{\rotatebox[origin=c]{90}{FANToM}} & 
        VNL  & $0.400_{\pm0.015}$ & $0.522_{\pm0.004}$ & $0.496_{\pm0.022}$ & $0.532_{\pm0.025}$ \\
        & CoT  & $0.398_{\pm0.014}$ & $0.516_{\pm0.011}$ & $0.482_{\pm0.030}$ & $0.521_{\pm0.024}$ \\
        & SToM & $0.413_{\pm0.012}$ & $0.537_{\pm0.027}$ & $0.379_{\pm0.024}$ & $0.516_{\pm0.014}$ \\
        & TToM & $0.252_{\pm0.020}$ & $0.476_{\pm0.019}$ & $0.419_{\pm0.018}$ & $0.409_{\pm0.026}$ \\
        & DWM & $0.429_{\pm0.013}$ & $0.558_{\pm0.012}$ & \underbar{$0.519_{\pm0.022}$} & $0.543_{\pm0.014}$ \\
        & PToM & $0.408_{\pm0.023}$ & \boldmath{$0.579_{\pm0.014}$} & \boldmath{$0.542_{\pm0.019}$} & \underbar{$0.573_{\pm0.016}$} \\
        \midrule
        \addlinespace[0.5ex]
        \rowcolor{tablewhite} & \textbf{\nkbp} & \underbar{$0.445_{\pm0.026}$} & $0.474_{\pm0.020}$ & $0.407_{\pm0.025}$ & $0.450_{\pm0.013}$ \\
        \rowcolor{tablewhite} & \textbf{\nkbt} & \boldmath{$0.487_{\pm0.018}$} & \underbar{$0.560_{\pm0.030}$} & $0.405_{\pm0.026}$ & \boldmath{$0.574_{\pm0.031}$} \\
        \addlinespace[0.5ex]
        \bottomrule
    \end{tabular}
    }
    \caption{Results of the scaling experiments. This table shows the exact mean accuracy of each method on ToMi, HiToM, and FANToM with base LLMs of different sizes. The results shown in this table are used to generate Figure~\ref{fig:scaling_heatmap}.}
    \label{app:tab:scaling_results}
\end{table*}

\section{Detailed Results of High-Order ToM Reasoning with \ours}
\label{app:high_order_tom}

\begin{figure*} [ht!]
    \centering
    \includegraphics[width=\textwidth]{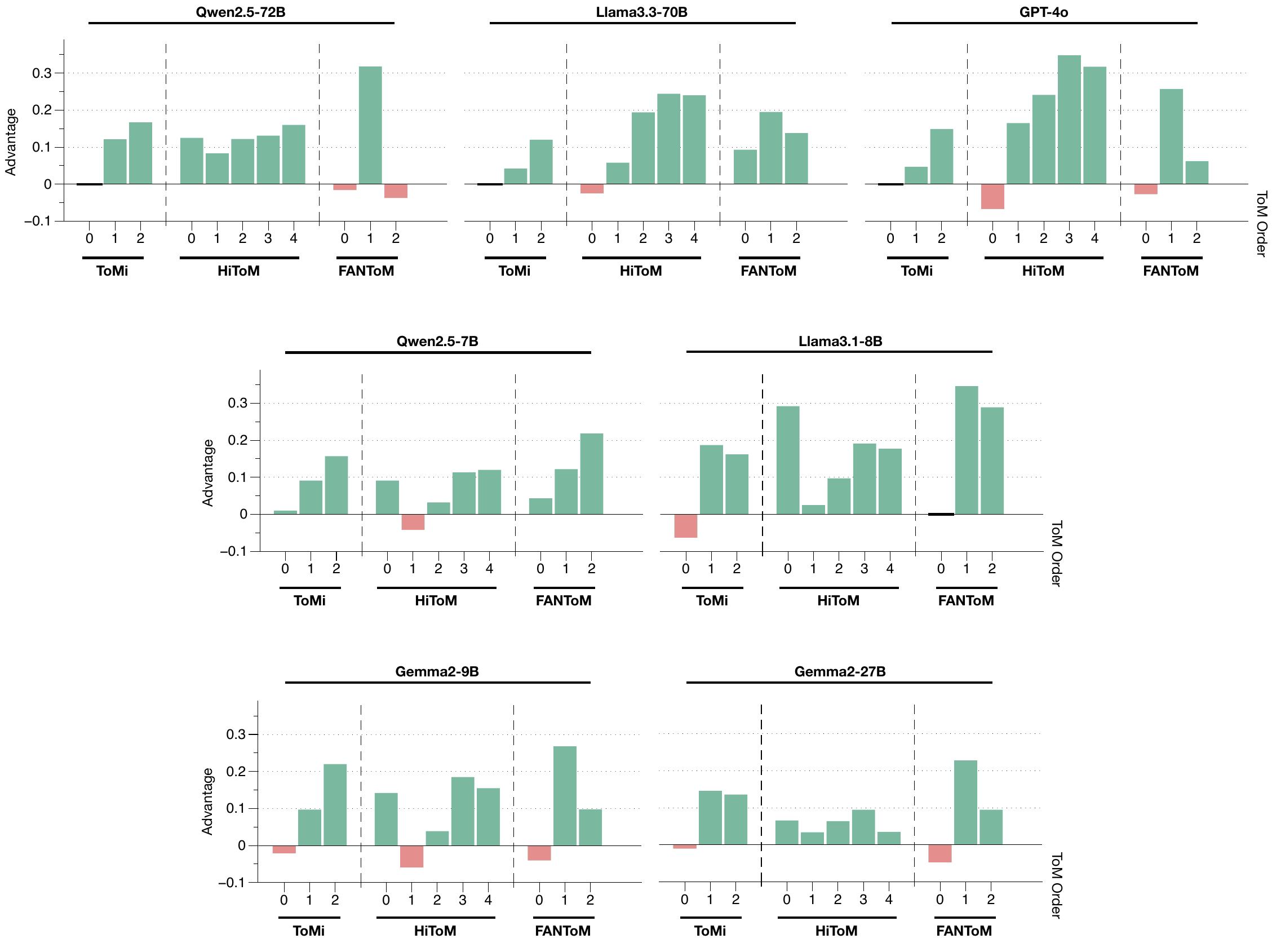}
    \caption{The complete results of the advantage of \ours~with respect the ToM order on ToMi, HiToM, and FANToM datasets.}
    \label{fig:high_order_all}
\end{figure*}

\paragraph{Complete Results}
As discussed in \S\ref{sec:analysis_high_order}, we observe that \ours~improves LLMs' capability of conducting high-order ToM reasoning using the HiToM dataset. Although the ToM reasoning order of ToMi and FANToM are limited to second-order, we observe a similar trend where \ours~brings consirable improvements in the case of second-order ToM reasoning. See Figure~\ref{fig:high_order_all} for detailed results.

\begin{table*} [ht!]
    \centering
    \large
    \resizebox{\linewidth}{!}{
    \begin{tabular} {P{0.01\columnwidth} c c c c c c c c c }
        \toprule
        & & \textbf{Qwen2.5-7B} & \textbf{Llama3.1-8B} & \textbf{Gemma2-9B} & \textbf{Gemma2-27B} & \textbf{Llama3.3-70B$^{\texttt{4bit}}$} & \textbf{Qwen2.5-72B$^{\texttt{4bit}}$} & \textbf{GPT-4o} \\
        \midrule
        \addlinespace[0.5ex]
        \multirow{4}{*}{\rotatebox[origin=c]{90}{ToMi}} & 
        \textbf{\nkbp} & $0.706_{\pm0.044}$ & $0.738_{\pm0.056}$ & $0.865_{\pm0.031}$ & $0.833_{\pm0.018}$ & $0.828_{\pm0.012}$ & $0.839_{\pm0.014}$ & $0.847_{\pm0.030}$ \\
        & \textbf{\nkbt} & $0.825_{\pm0.030}$ & $0.796_{\pm0.023}$ & $0.814_{\pm0.020}$ & $0.804_{\pm0.050}$ & $0.787_{\pm0.024}$ & $0.837_{\pm0.024}$ & $0.795_{\pm0.036}$ \\
        & $\quad\text{w/o }\texttt{KI}$ & \tabworse{$0.805_{\pm0.018}$} & \tabsame{$0.799_{\pm0.022}$} & \tabworse{$0.842_{\pm0.023}$} & \tabworse{$0.825_{\pm0.031}$} & \tabbetter{$0.834_{\pm0.067}$} & \tabbetter{$0.845_{\pm0.026}$} & \tabworse{$0.811_{\pm0.028}$} \\
        & $\quad\text{w/o }\texttt{IM}$ & \tabworse{$0.623_{\pm0.021}$} & \tabworse{$0.597_{\pm0.040}$} & \tabworse{$0.686_{\pm0.019}$} & \tabworse{$0.658_{\pm0.027}$} & \tabworse{$0.693_{\pm0.014}$} & \tabworse{$0.655_{\pm0.039}$} & \tabworse{$0.674_{\pm0.002}$} \\
        \addlinespace[0.5ex]
        \bottomrule
        \addlinespace[0.5ex]
        \multirow{4}{*}{\rotatebox[origin=c]{90}{HiToM}} & 
        \textbf{\nkbp} & $0.508_{\pm0.012}$ & $0.477_{\pm0.005}$ & $0.555_{\pm0.010}$ & $0.576_{\pm0.004}$ & $0.696_{\pm0.007}$ & $0.605_{\pm0.007}$ & $0.733_{\pm0.017}$ \\
        & \textbf{\nkbt} & $0.457_{\pm0.005}$ & $0.431_{\pm0.010}$ & $0.446_{\pm0.008}$ & $0.478_{\pm0.004}$ & $0.518_{\pm0.011}$ & $0.473_{\pm0.010}$ & $0.626_{\pm0.020}$ \\
        & $\quad\text{w/o }\texttt{KI}$ & \tabbetter{$0.511_{\pm0.007}$} & \tabworse{$0.463_{\pm0.006}$} & \tabbetter{$0.571_{\pm0.012}$} & \tabbetter{$0.614_{\pm0.007}$} & \tabbetter{$0.726_{\pm0.004}$} & \tabbetter{$0.632_{\pm0.003}$} & \tabbetter{$0.751_{\pm0.004}$} \\
        & $\quad\text{w/o }\texttt{IM}$ & \tabworse{$0.380_{\pm0.004}$} & \tabworse{$0.341_{\pm0.015}$} & \tabworse{$0.406_{\pm0.006}$} & \tabworse{$0.408_{\pm0.013}$} & \tabworse{$0.460_{\pm0.013}$} & \tabworse{$0.423_{\pm0.008}$} & \tabworse{$0.442_{\pm0.006}$} \\
        \addlinespace[0.5ex]
        \bottomrule
        \addlinespace[0.5ex]
        \multirow{4}{*}{\rotatebox[origin=c]{90}{FANToM}} & 
        \textbf{\nkbp} & $0.445_{\pm0.026}$ & $0.442_{\pm0.018}$ & $0.439_{\pm0.023}$ & $0.462_{\pm0.014}$ & $0.515_{\pm0.020}$ & $0.450_{\pm0.013}$ & $0.531_{\pm0.015}$ \\
        & \textbf{\nkbt} & $0.487_{\pm0.018}$ & $0.545_{\pm0.036}$ & $0.530_{\pm0.012}$ & $0.582_{\pm0.028}$ & $0.610_{\pm0.021}$ & $0.574_{\pm0.031}$ & $0.553_{\pm0.011}$\\
        & $\quad\text{w/o }\texttt{KI}$ & \tabworse{$0.487_{\pm0.022}$} & \tabsame{$0.544_{\pm0.033}$} & \tabsame{$0.530_{\pm0.017}$} & \tabworse{$0.579_{\pm0.021}$} & \tabworse{$0.607_{\pm0.018}$} & \tabworse{$0.542_{\pm0.036}$} & \tabworse{$0.539_{\pm0.012}$}\\
        & $\quad\text{w/o }\texttt{IM}$ & \tabworse{$0.436_{\pm0.028}$} & \tabworse{$0.448_{\pm0.024}$} & \tabworse{$0.426_{\pm0.011}$} & \tabworse{$0.478_{\pm0.008}$} & \tabworse{$0.500_{\pm0.021}$} & \tabworse{$0.477_{\pm0.017}$} & \tabworse{$0.470_{\pm0.013}$} \\
        \addlinespace[0.5ex]
        \bottomrule
    \end{tabular}
    }
    \caption{Results of the ablation study of \ours~on ToMi, HiToM, and FANToM datasets. "w/o \texttt{KI}" indicates without \textit{entity state knowledge injection}. "w/o \texttt{IM}" denotes without \textit{perspective-taking via iterative masking}.}
    \label{app:tab:ablation_results}
\end{table*}

\paragraph{Ablation Study}
In \S\ref{sec:analysis_ablation}, we analyzed the efficacy of the \texttt{IM} and \texttt{KI} components of \ours. Here, we conduct analysis on the impact of \texttt{IM} and \texttt{KI} with respect to the ToM reasoning order. Specifically, we compute the advantage of removing the \texttt{IM} or \texttt{KI} component from \ours~as
\begin{equation}
\text{Adv} = \text{Acc}_{\text{w/o component}} - \text{Acc}_{\text{w/ component}}
\label{app:eq:advantage}
\end{equation}
where a positive advantage score means that the model performed better without the component and vice versa.
Table~\ref{app:tab:ablation_high_order_tomi}, Table~\ref{app:tab:ablation_high_order_hitom}, and Table~\ref{app:tab:ablation_high_order_fantom} shows the result of the ablation study. Aligning with our findings in \S\ref{sec:analysis_ablation}, \texttt{IM} is indispensable to all ToM reasoning tasks, reiterating the importance of perspective-taking in ToM reasoning. On the other hand, \texttt{KI} is critical for the dialogue-based FANToM dataset while less effective for the event-based ToMi and HiToM datasets. This finding illustrates that \nkb's functionality as an information compressor by compressing key information as entity state knowledge is substantial when tackling ToM reasoning with sparse information (as in daily dialogue). Further, removing \texttt{KI} leads to larger performance degradation in LLMs of smaller size, further highlighting that small LLMs are less capable of dealing with reporting bias and prefer to have explicit state information regarding key entities in events. 

\section{Detailed Results of Scaling Experiment}
The exact mean accuracies of \ours~and baseline methods on ToMi, HiToM, and FANToM datasets using LLMs of varying sizes are shown in Table~\ref{app:tab:scaling_results}. 

\begin{table*} [ht!]
    \centering
    \resizebox{\linewidth}{!}{
    \begin{tabular} {P{0.01\columnwidth} c c c c c c c c c }
        \toprule
        & ToM Order & \textbf{Qwen2.5-7B} & \textbf{Llama3.1-8B} & \textbf{Gemma2-9B} & \textbf{Gemma2-27B} & \textbf{Llama3.3-70B$^{\texttt{4bit}}$} & \textbf{Qwen2.5-72B$^{\texttt{4bit}}$} & \textbf{GPT-4o} \\
        \midrule
        \addlinespace[0.5ex]
        \multirow{3}{*}{\rotatebox[origin=c]{90}{w/o \texttt{KI}}} 
        & \textit{Zeroth} & $+0.008$ & $+0.016$ & $+0.008$ & $-0.000$ & $+0.008$ & $-0.000$ & $-0.000$ \\
        & \textit{First} & $-0.019$ & $-0.019$ & $-0.000$ & $+0.005$ & $+0.014$ & $-0.000$ & $-0.060$ \\
        & \textit{Second} & $-0.004$ & $-0.004$ & $-0.039$ & $-0.022$ & $+0.017$ & $-0.025$ & $-0.030$\\
        
        \addlinespace[0.5ex]
        \midrule
        \addlinespace[0.5ex]
        
        \multirow{3}{*}{\rotatebox[origin=c]{90}{w/o \texttt{IM}}} 
        & \textit{Zeroth} & $+0.008$ & $-0.024$ & $-0.000$ & $-0.000$ & $+0.008$ & $-0.000$ & $+0.000$ \\
        & \textit{First} & $-0.127$ & $-0.226$ & $-0.161$ & $-0.283$ & $-0.273$ & $-0.025$ & $-0.164$ \\
        & \textit{Second} & $-0.025$ & $-0.124$ & $-0.253$ & $-0.224$ & $-0.206$ & $-0.218$ & $-0.269$ \\
        \bottomrule
    \end{tabular}
    }
    \caption{Results of the ablation study of \ours~on ToMi with respect to ToM order. We display the advantage scores computed using Equation~\ref{app:eq:advantage}. "w/o \texttt{KI}" indicates without \textit{entity state knowledge injection}. "w/o \texttt{IM}" denotes without \textit{perspective-taking via iterative masking}.}
    \label{app:tab:ablation_high_order_tomi}
\end{table*}

\begin{table*} [ht!]
    \centering
    \resizebox{\linewidth}{!}{
    \begin{tabular} {P{0.01\columnwidth} c c c c c c c c c }
        \toprule
        & ToM Order & \textbf{Qwen2.5-7B} & \textbf{Llama3.1-8B} & \textbf{Gemma2-9B} & \textbf{Gemma2-27B} & \textbf{Llama3.3-70B$^{\texttt{4bit}}$} & \textbf{Qwen2.5-72B$^{\texttt{4bit}}$} & \textbf{GPT-4o} \\
        \midrule
        \addlinespace[0.5ex]
        \multirow{5}{*}{\rotatebox[origin=c]{90}{w/o \texttt{KI}}} 
        & \textit{Zeroth} & $-0.017$ & $+0.075$ & $+0.058$ & $+0.005$ & $+0.058$ & $+0.042$ & $+0.075$ \\
        & \textit{First} & $-0.008$ & $-0.017$ & $+0.008$ & $-0.017$ & $+0.016$ & $-0.017$ & $+0.023$ \\
        & \textit{Second} & $+0.045$ & $+0.039$ & $+0.026$ & $+0.013$ & $+0.019$ & $+0.006$ & $+0.002$ \\
        & \textit{Third} & $-0.018$ & $+0.018$ & $+0.018$ & $+0.065$ & $+0.054$ & $+0.030$ & $0.009$ \\
        & \textit{Fourth} & $+0.040$ & $+0.040$ & $+0.017$ & $+0.046$ & $+0.040$ & $+0.029$ & $+0.008$ \\
        
        \addlinespace[0.5ex]
        \midrule
        \addlinespace[0.5ex]
        
        \multirow{5}{*}{\rotatebox[origin=c]{90}{w/o \texttt{IM}}} 
        & \textit{Zeroth} & $-0.000$ & $-0.000$ & $-0.000$ & $-0.000$ & $-0.000$ & $-0.008$ & $+0.016$ \\
        & \textit{First} & $-0.019$ & $-0.100$ & $-0.092$ & $-0.150$ & $-0.259$ & $-0.175$ & $-0.229$ \\
        & \textit{Second} & $-0.142$ & $-0.084$ & $-0.207$ & $-0.193$ & $-0.323$ & $-0.271$ & $-0.403$ \\
        & \textit{Third} & $-0.214$ & $-0.143$ & $-0.238$ & $-0.232$ & $-0.268$ & $-0.261$ & $-0.365$ \\
        & \textit{Fourth} & $-0.125$ & $-0.103$ & $-0.211$ & $-0.205$ & $-0.257$ & $-0.177$ & $-0.316$ \\
        \bottomrule
    \end{tabular}
    }
    \caption{Results of the ablation study of \ours~on HiToM with respect to ToM order. We display the advantage scores computed using Equation~\ref{app:eq:advantage}. "w/o \texttt{KI}" indicates without \textit{entity state knowledge injection}. "w/o \texttt{IM}" denotes without \textit{perspective-taking via iterative masking}.}
    \label{app:tab:ablation_high_order_hitom}
\end{table*}

\begin{table*} [ht!]
    \centering
    \resizebox{\linewidth}{!}{
    \begin{tabular} {P{0.01\columnwidth} c c c c c c c c c }
        \toprule
        & ToM Order & \textbf{Qwen2.5-7B} & \textbf{Llama3.1-8B} & \textbf{Gemma2-9B} & \textbf{Gemma2-27B} & \textbf{Llama3.3-70B$^{\texttt{4bit}}$} & \textbf{Qwen2.5-72B$^{\texttt{4bit}}$} & \textbf{GPT-4o} \\
        \midrule
        \addlinespace[0.5ex]
        \multirow{3}{*}{\rotatebox[origin=c]{90}{w/o \texttt{KI}}} 
        & \textit{Zeroth} & $-0.032$ & $-0.054$ & $-0.031$ & $-0.044$ & $-0.053$ & $-0.058$ & $-0.010$ \\
        & \textit{First} & $-0.308$ & $-0.255$ & $-0.265$ & $-0.392$ & $-0.237$ & $-0.205$ & $+0.005$ \\ 
        & \textit{Second} & $-0.162$ & $-0.212$ & $-0.113$ & $-0.230$ & $-0.128$ & $-0.119$ & $-0.005$ \\
        
        \addlinespace[0.5ex]
        \midrule
        \addlinespace[0.5ex]
        
        \multirow{3}{*}{\rotatebox[origin=c]{90}{w/o \texttt{IM}}} 
        & \textit{Zeroth} & $-0.012$ & $-0.003$ & $-0.025$ & $-0.027$ & $-0.041$ & $-0.035$ & $-0.025$ \\
        & \textit{First} & $-0.335$ & $-0.367$ & $-0.343$ & $-0.370$ & $-0.327$ & $-0.418$ & $-0.348$ \\
        & \textit{Second} & $-0.178$ & $-0.215$ & $-0.126$ & $-0.218$ & $-0.144$ & $-0.073$ & $+0.000$ \\
        \bottomrule
    \end{tabular}
    }
    \caption{Results of the ablation study of \ours~on FANToM with respect to ToM order. We display the advantage scores computed using Equation~\ref{app:eq:advantage}. "w/o \texttt{KI}" indicates without \textit{entity state knowledge injection}. "w/o \texttt{IM}" denotes without \textit{perspective-taking via iterative masking}.}
    \label{app:tab:ablation_high_order_fantom}
\end{table*}

\section{Examples of Knowledge Injection}
\label{app:ki_examples}
We demonstrate how \texttt{KI} is accomplished with an event sequence from ToMi. The important entities and attributes are generated using Llama3.3-70B$^\texttt{4bit}$ and the entity state knowledge is generated by \nkbt$_\texttt{8B}$.

\begin{tcolorbox}[
    enhanced,
    attach boxed title to top center={
        yshift=-3mm,yshifttext=-1mm
    },
    breakable,
    colframe=pink!90!black,
    colbacktitle=pink,
    title=Events from ToMi,
    coltitle=black,
    collower=black, 
    label=tomi_ki_example_part1,
    fonttitle=\bfseries,
    boxed title style={size=small,colframe=pink!75!black}
    ]
    1: Benjamin entered the crawlspace. \\
    2: Abigail entered the crawlspace. \\
    3: Emily entered the crawlspace. \\
    4: The t-shirt is in the cupboard. \\
    5: The cupboard is in the crawlspace. \\
    6: Abigail exited the crawlspace. \\
    7: Emily moved the t-shirt to the basket. \\
    8: The basket is in the crawlspace. \\
    9: Benjamin hates the coat. \\
    10: Emily exited the crawlspace. \\
    11: Abigail entered the crawlspace.
\end{tcolorbox}

\begin{tcolorbox}[
    enhanced,
    attach boxed title to top center={
        yshift=-3mm,yshifttext=-1mm
    },
    breakable,
    colframe=pink!90!black,
    colbacktitle=pink,
    title=Entity State Generated by \nkbt$_\texttt{8B}$,
    coltitle=black,
    collower=black, 
    label=tomi_ki_example_part2,
    fonttitle=\bfseries,
    boxed title style={size=small,colframe=pink!75!black}
    ]
    \textbf{Event1:} \\
    - Location of Benjamin is in the crawlspace. \\

    \textbf{Event2:} \\
    - Location of Abigial is in the crawlspace. \\

    \textbf{Event3:} \\
    - Location of Emily is in the crawlspace. \\

    \textbf{Event4:} \\
    - Location of T-shirt is in cupboard. \\

    \textbf{Event5:} \\
    - Location of cupboard is in crawlspace. \\

    \textbf{Event6:} \\
    - Location of Abigail is outside the crawlspace. \\

    \textbf{Event7:} \\
    - Location of T-shirt is in basket. \\
    - Content of basket is T-shirt. \\
    - Content of cupboard is empty. \\

    \textbf{Event8:} \\
    - Location of basket is in the crawlspace. \\

    \textbf{Event9:} \\
    None

    \textbf{Event10:} \\
    - Location of Emily is outside the crawlspace. \\

    \textbf{Event11:} \\
    - Location of Abigail is in the crawlspace.
    
\end{tcolorbox}

\begin{tcolorbox}[
    enhanced,
    attach boxed title to top center={
        yshift=-3mm,yshifttext=-1mm
    },
    breakable,
    colframe=pink!90!black,
    colbacktitle=pink,
    title=Augmented ToMi Events,
    coltitle=black,
    collower=black, 
    label=tomi_ki_example_part4,
    fonttitle=\bfseries,
    boxed title style={size=small,colframe=pink!75!black}
    ]
    1: Benjamin entered the crawlspace. \\

    2: Abigail entered the crawlspace. \\

    3: Emily entered the crawlspace. \\

    4: The t-shirt is in the cupboard. \\
    - Location of T-shirt is in the cupboard. \\

    5: The cupboard is in the crawlspace. \\
    - Location of cupboard is in the crawlspace. \\

    6: Abigail exited the crawlspace. \\

    7: Emily moved the t-shirt to the basket. \\
    - Location of T-shirt is in basket. 
    - Content of basket is T-shirt. \\
    - Content of cupboard is empty. \\

    8: The basket is in the crawlspace. \\
    - Location of basket is in the crawlspace. \\

    9: Benjamin hates the coat. \\

    10: Emily exited the crawlspace. \\

    11: Abigail entered the crawlspace.
\end{tcolorbox}

\section{Examples of Entity State Knowledge Generated by \nkbt$_\texttt{8B}$ and \nkbt$_\texttt{70B}$}
\label{app:nkbt_comparisons}

\paragraph{\textit{Entity states generated by \nkbt$_{\texttt{70B}}$ is more informative compared to \nkbt$_\texttt{8B}$.}}

From the annotated entity states generated by \nkbt$_\texttt{8B}$ and \nkbt$_\texttt{70B}$ in both ToMi and FANToM, we observe that \nkbt$_\texttt{70B}$ is much more eloquent compared to its counterparts. Such eloquence makes the entity state knowledge more informative in ToMi:

\begin{tcolorbox}[
    enhanced,
    attach boxed title to top center={
        yshift=-3mm,yshifttext=-1mm
    },
    breakable,
    colframe=purple!75!black,
    colbacktitle=purple,
    title=Entity State for ToMi,
    coltitle=white2,
    collower=black, 
    label=tomi_es_informative,
    fonttitle=\bfseries,
    boxed title style={size=small,colframe=purple!75!black}
    ]
        \textbf{Example Event 1}  \\
        Amelia moved the belt to the pantry. \\

        \textbf{\nkbt$_\texttt{8B}$}  \\
        Amelia's knowledge of the belt's location is \textcolor{green2}{\underline{known}} \\

        \textbf{\nkbt$_\texttt{70B}$}  \\
        Amelia's knowledge of the belt's location is \textcolor{green2}{\underline{that the belt is in the pantry}}
        \tcblower

        \textbf{Example Event 2} \\
        Lucas exited the living room. \\

        \textbf{\nkbt$_\texttt{8B}$}  \\
        Location of Lucas as known to Charlotte is \textcolor{green2}{\underline{absent}}

        \textbf{\nkbt$_\texttt{70B}$}  \\
        Location of Lucas as known to Charlotte is \textcolor{green2}{\underline{outside the living room}}

\end{tcolorbox}

However, in FANToM, the increased eloquence of \nkbt$_\texttt{70B}$ leads to more hallucination, resulting in more erroneous information and reducing its functionality as an information compressor (contents that can be entailed from the utterance is marked in \improved~green whereas the irrelevant contents are marked in \worse~red):

\begin{tcolorbox}[
    enhanced,
    attach boxed title to top center={
        yshift=-3mm,yshifttext=-1mm
    },
    breakable,
    colframe=purple!75!black,
    colbacktitle=purple,
    title=Entity State for FANToM,
    coltitle=white2,
    collower=black, 
    label=tomi_es_informative,
    fonttitle=\bfseries,
    boxed title style={size=small,colframe=yellow!50!black}
    ]
        \textbf{Example Event 1}  \\
        Breanna: Long-distance, yes, but not for as long as you both. It was a few months, and things didn't work out for us. But talking about relationships, it brings back memories of my first crush.

        \textbf{\nkbt$_\texttt{8B}$}  \\
        Breanna's knowledge of longdistance relationship methods is \textcolor{green2}{\underline{known}}.

        \textbf{\nkbt$_\texttt{70B}$}  \\
        Breanna's knowledge of longdistance relationship methods is \textcolor{red}{\ul{limited to}}\textcolor{green2}{\ul{ her past personal experience of a few months,}}\textcolor{green2}{\ul{ with an unsuccessful outcome,}}\textcolor{red}{\ul{ and she has been exposed to Nickolas and Raelynn's methods, including the importance of trust, understanding, commitment, and regular communication, such as scheduling regular talks and quick calls.}}
        \tcblower

        \textbf{Example Event 2} \\
        Shayla: It was. But I made it through. It was a journey that brought me face-to-face with my own strength. It made me realize the importance of mental health and inspired me to help others who are going through something similar. \\

        \textbf{\nkbt$_\texttt{8B}$}  \\
        Significance of nurturing mental wellbeing is \color{green2}{\ul{important}}.

        \textbf{\nkbt$_\texttt{70B}$}  \\
        Significance of nurturing mental wellbeing is \color{green2}{\ul{heightened}}\color{red}{\ul{, as it is recognized as a crucial aspect of overall health, particularly in the aftermath of traumatic experiences like abusive relationships, and is seen as a key factor in building resilience and facilitating recovery}}.
\end{tcolorbox}

\section{Evaluation of \nkbsectiontitle}
\label{app:annotation}
As discussed in \S\ref{sec:analysis_scaling} and Table~\ref{tab:nkbt_annotation}, we manually labeled 300 instances of entity state information from ToMi and FANToM datasets. The annotation was carried out by three graduate students (100 annotations each) majoring in computer science at a prestigious university. Since the annotation was conducted for entity state information, which only contains commonsense knowledge, no training was provided to the annotators. An annotation guideline was provided to the annotators containing a description of the annotation task and examples that cover all possible label combinations. 

The data provided to the annotators are identical to the ones used in \ours. The set of key entities ($E$) and their attributes ($A$) are generated using Llama3.3-70B$^{\texttt{4bit}}$. Hence, the \texttt{relevance score} reflects Llama3.3-70B$^{\texttt{4bit}}$'s capability of generating entities and attributes that are important to answering the corresponding ToM questions. Although it is not shown in the example, we provide the complete set of questions and cumulative events for each data point. In other words, given an event sequence $\{\epsilon\}_{i=1, n}$ and a set of ToM questions $\mathcal{Q}$, we provided $\epsilon_{1:k}$ and $\mathcal{Q}$ when annotating the $k^{th}$ event and its entity states. 

The detailed annotation guideline is shown in the following pages. 

\newpage
\onecolumn
\section*{\centering \Huge Annotation Guideline for Evaluation of Neural Knowledge Base}

\vspace{1.5em}
\section*{Task Formulation}

In this task, you will be responsible for evaluating the quality of entity state information generated by two Neural Knowledge Base. You will be given a set of \textbf{Questions} in the \texttt{question} column, an \textbf{event} or \textbf{utterance} in the \texttt{event} column and a \textbf{entity state information} generated from a neural knowledge bank in the \texttt{XXX\_states} column. 

\vspace{1.5em}
\noindent
You will be annotating the following two aspects of the generated entity state information:
\begin{itemize}[leftmargin=6mm, noitemsep]
    \item[1.] \textbf{Relevance}: whether the generated entity state knowledge is helpful with answering the questions. \\
    \item[2.] \textbf{Accuracy}: whether the entity state can be entailed from the given event/utterance.
\end{itemize}

For both tasks, choose \textbf{Yes} if you believe that the entity state is relevant/accurate and choose \textbf{No} otherwise. 

Note that all of the entity state follows a semi-structured formulation:
\begin{quote}
\tt - \textbf{[Attribute]} of \textbf{[Entity]} is \textbf{[Current State]}
\end{quote}

In the \textbf{Relevance} annotation task, you only need to pay attention to the \texttt{\textbf{[Attribute] of [Entity]}}.

In the \textbf{Accuracy} annotation task, you need to consider the complete entity state description. 

\vspace{1.5em}
\section*{Examples}
\noindent
\noindent\rule{16cm}{2pt} \\
\textbf{Question}: Where does Abigail think that Benjamin searches for the pumpkin? \\
\textbf{Event}: Abigail moved the pumpkin to the bathtub. \\
\textbf{State}: \textit{location} of \textit{the pumpkin} is \textit{in bathtub}

\vspace{1em}
\noindent
In this example, we see that the \texttt{Location of the Pumpkin} is important in answering the question. \textbf{Therefore, we mark the relevance as \texttt{Yes}.}
We see that there exists an entailment relationship between \texttt{Event} and \texttt{State}. \textbf{Therefore, we mark the accuracy as \texttt{Yes}}.

\vspace{2em}
\noindent\rule{16cm}{2pt} \\
\textbf{Question}: Where does Abigail think that Benjamin searches for the pumpkin? \\
\textbf{Event}: Chloe entered the workshop. \\
\textbf{State}: \textit{location} of \textit{Chloe} is \textit{in workshop}

\vspace{1em}
\noindent
In this example, we see that the \texttt{Location of the Chloe} is not related to the question. \textbf{Therefore, we mark the relevance as \texttt{No}.}

We see that there exists an entailment relationship between \texttt{Event} and \texttt{State}. \textbf{Therefore, we mark the accuracy as \texttt{Yes}.}

\vspace{2em}
\noindent\rule{16cm}{2pt} \\
\textbf{Question}: Where does Abigail think that Benjamin searches for the pumpkin? \\
\textbf{Event}: Abigail moved the pumpkin to the bathtub. \\
\textbf{State}: \textit{location} of \textit{Abigail} is \textit{in bathtub}

\vspace{1em}
\noindent
In this example, we see that the \texttt{Location of the Abigail} is important in answering the question. \textbf{Therefore, we mark the relevance as \texttt{Yes}.}

However, it is unlikely that Abigail needs to enter the bathtub to place the pumpkin. \textbf{Therefore, there does not exist an entailment relationship between \texttt{Event} and \texttt{State}. Therefore, we mark the accuracy as \texttt{No}.}

\vspace{2em}
\noindent\rule{16cm}{2pt} \\
\textbf{Question}: Where does Abigail think that Benjamin searches for the pumpkin? \\
\textbf{Event}: Chloe entered the workshop. \\
\textbf{State}: \textit{location} of \textit{Chloe} is \textit{in kitchen}

\vspace{1em}
\noindent
In this example, we see that the \texttt{Location of the Chloe} is not related to the question. \textbf{Therefore, we mark the relevance as \texttt{No}.}

We see that there does not exist an entailment relationship between \texttt{Event} and \texttt{State}. \textbf{Therefore, we mark the accuracy as \texttt{No}.} \\ 
\noindent\rule{16cm}{2pt}

\end{document}